\documentclass{article} 
\usepackage{iclr2027_conference,times}
\usepackage{algorithm}
\usepackage{algpseudocode}
\usepackage{makecell}
 \usepackage{multirow}

\usepackage{amsmath,amsfonts,bm}

\def\eqref#1{equation~\ref{#1}}

\def\1{\bm{1}}

\DeclareMathAlphabet{\mathsfit}{\encodingdefault}{\sfdefault}{m}{sl}
\SetMathAlphabet{\mathsfit}{bold}{\encodingdefault}{\sfdefault}{bx}{n}

\definecolor{nicepurple}{RGB}{148,0,211}

\usepackage{url}
\usepackage{graphicx}

\usepackage[colorlinks = true,
            linkcolor = black,
            urlcolor  = blue,
            citecolor = teal,
            anchorcolor = blue]{hyperref}

\usepackage{enumitem}

\usepackage{wrapfig,bm,amsmath}  
\usepackage{booktabs}
\usepackage{etoc}
\usepackage{wrapfig}

\title{Evolving Dexterous Robots from Scratch}

\author{%
Zihan Guo,$^*$ Shuzhe Zhang,$^*$ Muhan Li,$^*$ Peiyang Li, Sam Kriegman
\\Northwestern University
}

\usepackage[table]{xcolor}
\definecolor{editcolor}{RGB}{205, 85, 0}

\definecolor{bestcol}{HTML}{9FCBF5}
\definecolor{secondcol}{HTML}{C8E1F9}
\newcommand{\best}{\cellcolor{bestcol}}
\newcommand{\second}{\cellcolor{secondcol}}

\iclrfinalcopy 
\begin{document}

\maketitle

\etocdepthtag.toc{mtchapter}

\begin{abstract}
Little is known about how to manually design agents capable of dexterous manipulation. 
Some design principles have been inferred from 
close examination of how animals
manipulate objects, 
but these structures and behaviors
have so far resisted biomimicry 
and 
may not be optimal for 
artificial machines.
Here we evolve freeform robots to pick up, hold, rotate, and use diverse objects.
Unlike other approaches to optimizing robot hands, 
we do not presuppose the presence, articulation, or geometry of any part of the body.
Although familiar prehensile forms such as tails, beaks, paws and claws
may emerge spontaneously under certain conditions---%
and while such conditions could be of interest to evolutionary biologists---%
de novo manipulator design can also
reveal whole new solutions,
overlooked or unknown structures
which may be better suited for the task at hand.
We use contrastive learning
to create a 
highly searchable
genetic embedding of design space,
an autoregressive 
developmental model to decode designs,
evolutionary strategies to find good designs, 
and reinforcement learning to train each evolved design. 
Winning designs were automatically converted into a manufacturable blueprint, printed, assembled and tested in the real world in a zero-shot manner.
The results represent the state-of-the-art in evolutionary robotics in terms of performance, diversity and complexity.
%

%
%
\end{abstract}

\begin{figure*}[!t]
  \centering
  \includegraphics[width=\columnwidth]{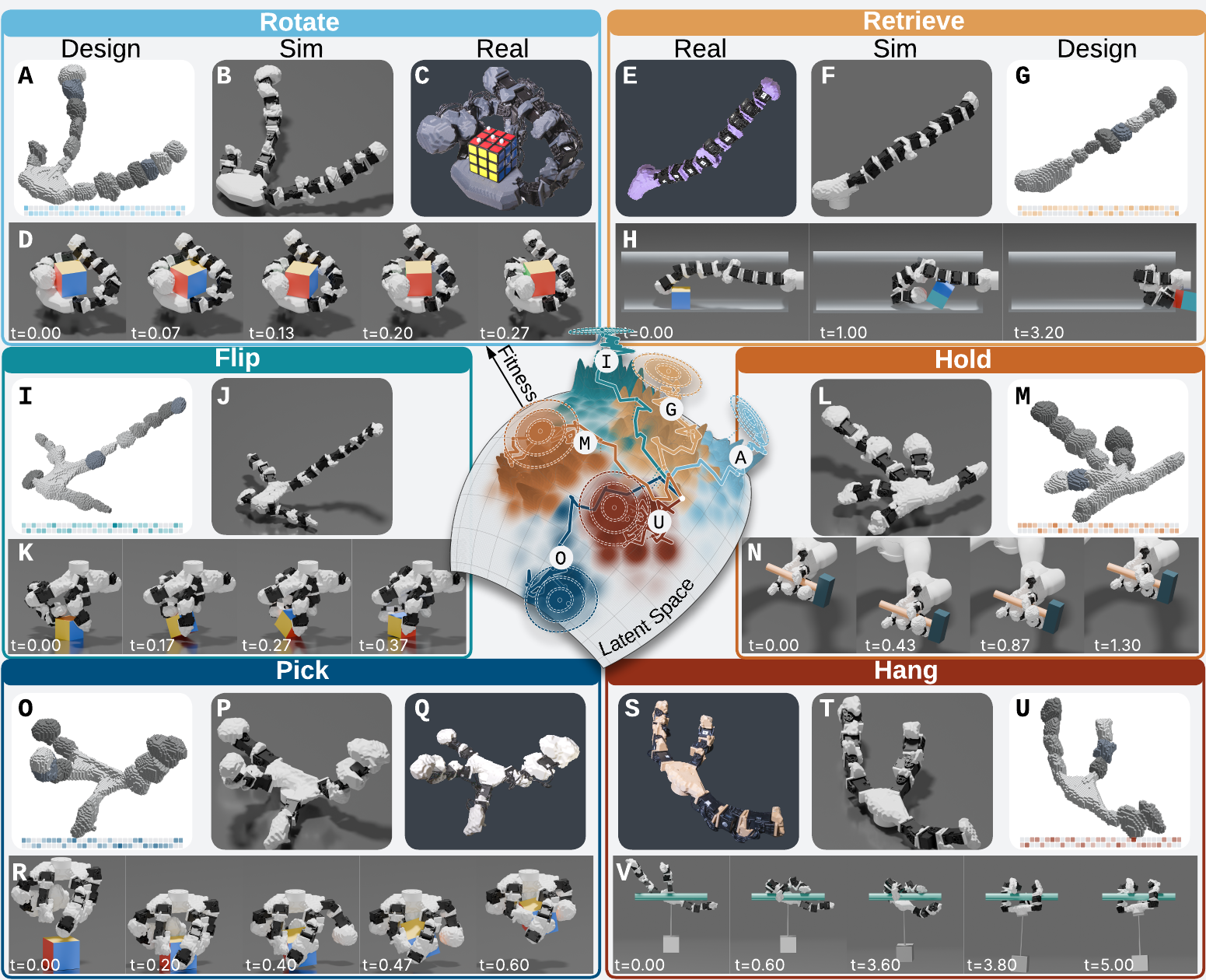}
  \vspace{-16pt} 
  \caption{\textbf{Evolving freeform manipulators.}
  Under different selection pressures and environmental conditions,
  unique morphologies emerged 
  to
  rotate (\textbf{A-D}),
  retrieve (\textbf{E-H}),
  flip (\textbf{I-K}),
  hold (\textbf{L-N}),
  pick up (\textbf{O-R}),
  and
  hang from (\textbf{S-V}) objects.
  Evolution operates in a latent space
  (multicolored surface, center inset)
  of genotypes, which are visualized as 
  colored barcodes below each decoded design (A,G,I,M,O,U).
  The resulting robot was 
  trained in simulation (D,H,K,N,R,V), and
  in some cases 
  built and tested in the real world (C,E,Q,S).
  }
  \vspace{-1em}
  \label{fig:evolution}
\end{figure*}

\section{Introduction}
\label{sec:intro}

A diversity of prehensile structures have evolved in animals to support effective object manipulation \citep{sugasawa2021object}.
While the specific solutions found in nature (e.g.~an elephant's trunk, an insect's mandibles, a primate's hand) might not be suitable for artificial machines, which are made of very different components and coordinated by very different control systems,
it may still be beneficial to mimic the evolutionary mechanisms responsible for the emergence, complexification, and diversification of nature's designs \citep{nolfi2002synthesis}.
That is, it can be useful and interesting to design robots automatically with minimal assumptions.
However, co-optimizing a robot's physical layout and control policy
is a notoriously challenging problem 
\citep{cheney2018scalable,strgar2024evolution,strgar2026accelerated}.

Several approaches for automatic robot design
have been reported in the literature.
The majority 
focused on locomotor performance 
and completely 
ignored object manipulation
\citep{lipson2000automatic,hiller2012automatic,brodbeck2015morphological,kriegman2020xenobots,matthews2023efficient,yu2026reconfigurable}.
However, some manipulator structures have also been optimized.
Usually
the size and shape of given links are deformed within a manually determined kinematic structure, 
such as a serial chain \citep{Xu-RSS-21}
or parallel jaw gripper \citep{ha2020fit2form,xu2025dynamics}.
Similarly,
\cite{liu2018optimal} optimized the 2D topology of compliant jaws,
and
\cite{bai2026learning} optimized 
the relative position and orientation of three soft fingers as well as the size, spacing, and tendon routing of four component blocks along each finger.
However, these underactuated grippers could only grasp and release,
and radically different anatomies (e.g.~more or less fingers) could not emerge. 

\cite{fay2026house} optimized the kinematic structure of a manipulator
using stacks of cubic modules to form inward-bending fingers of variable position and length.
The optimized design, when built, exhibited faster in-hand object rotation than an anthropomorphic baseline \citep{shaw2023leaphand}.
However, 
the shape and articulation of each finger was largely presupposed,
and their position was locked in plane about a 2D palm.
As a result, a limited diversity of designs could arise,
and 
differences between unoptimized and optimized manipulators
could be difficult to discern.
\cite{mirzaee2026function} likewise optimized the distribution and length of modular rectangular fingers
as well as the 3D surface curvature of the palm and the relative bend angle of each finger,
which improved grasp stability.
However, their anthropomorphic design template added more assumptions than it relaxed.

Here we explore 
the open-ended evolution of novel dexterous robots with freeform geometry and articulation (Fig.~\ref{fig:evolution}).
For fair comparison, we use the same actuators 
as the most relevant prior work \citep{fay2026house,mirzaee2026function}.
Unlike prior work however, 
we do not assume 
fingers about a palm
and do not constrain the position and orientation of actuators
or
the geometry of the segments between them.
After jointly co-optimizing the robot's controller and complex morphology from scratch in simulation,
we automatically convert the design into a manufacturable blueprint \citep{guo2026creating}, 
which we directly follow to print, assemble, and test the robot in the real world.

To support the evolution of
freeform manipulators, 
we encode the enormous, 
combinatorial space of 
possible morphologies---morphospace \citep{raup1966geometric}---%
into a compressed latent genome \citep{mitchell2025genomic}.
The related work mentioned above
\citep{fay2026house,mirzaee2026function}, by contrast,
operated within 
manually devised 
parameterizations of 
design space, 
as did other investigations of alternative hand designs \citep{gilday2025embodied}.
In locomotion studies
it is becoming increasingly popular
to
use a variational autoencoder 
(VAE; \cite{kingma2014auto})
to construct a latent embedding of design space \citep{hu2023glso,li2025generating,yu2026reconfigurable,wang2026ecomoe,guo2026creating}.
Similarly, \citet{wei2026one} used a VAE to operate kinematically distinct anthropomorphic hands with a unified control policy.
However, VAEs
can unintentionally obstruct evolution by mapping
nearly identical designs (phenotypes) to very distant latent codes (genotypes).
Here, 
we use contrastive learning
\citep{pmlr-v119-chen20j,pmlr-v119-wang20k,kirchoff2024salsa}
to 
project morphospace 
onto a genetic hypersphere,
which yields a phenotypically coherent---and more easily searchable---latent manifold.

\begin{figure*}[!t]
  \centering
  \includegraphics[width=\columnwidth]{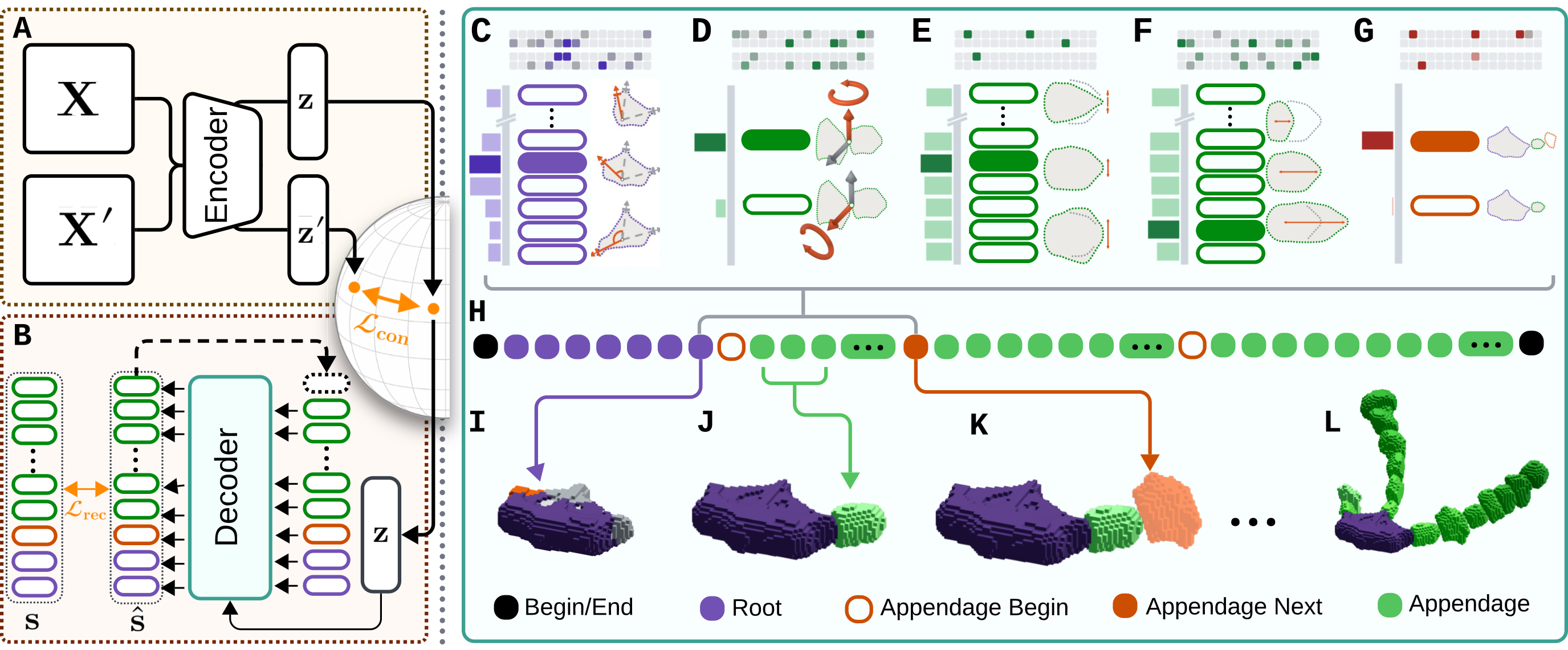}
  \vspace{-18pt} 
  \caption{\textbf{Latent genome and developmental decoder.}
  Morphologies were embedded on a latent manifold (\textbf{A}) and
  autoregressively decoded back to morphospace (\textbf{B}) 
  through a developmental program (\textbf{C-G})
  that translates design tokens (\textbf{H})
  into articulated bodies (\textbf{I-L}).
  %
  Dimensions of an evolved genotype (barcodes in C-G) most strongly correlated with each design token are highlighted in the corresponding color at each growth step. 
  The decoder outputs a probability distribution (vertical histogram in C) for each token that determines how new structures are layered on top of earlier structures, and their
  3D geometry (E,F) and articulation (D,G).
  }
  \vspace{-1em}
  \label{fig:development}
\end{figure*}

Another limitation of prior genetic embeddings has been the decoder, 
the genotype-to-phenotype map \citep{pigliucci2010genotype,mitchell2025genomic}.
This mapping was either too constrained, e.g.~limiting phenotypes to stick figures
\citep{hu2023glso,yu2026reconfigurable},
or too unconstrained, 
i.e.~with the capacity to produce nonsensical anatomies, such as those with floating, unconnected body parts \citep{li2025generating,guo2026creating}.
Here we introduce an autoregressive developmental model that 
iteratively grows one body part after another,
ensuring validity at each step.
It does so by
translating a sequence of design tokens (specified by the latent genotype) into morphological features, with different tokens determining how, when and where to layer new structures on top of older ones, 
and the shape and articulation of each part (Fig.~\ref{fig:development}).

\textbf{TL;DR:}
In short, here we show for the first time de novo manipulator design.
Key to our approach is a well-structured latent space 
for freeform morphological evolution
and a pipeline to instantly CADify decoded designs, obviating the need for standardized body modules.
The results are more diverse, complex, and dexterous
than other evolved robots reported in the literature to date.

\begin{figure*}[!t]
  \centering
  \vspace{-1em} 
  \includegraphics[width=\columnwidth]{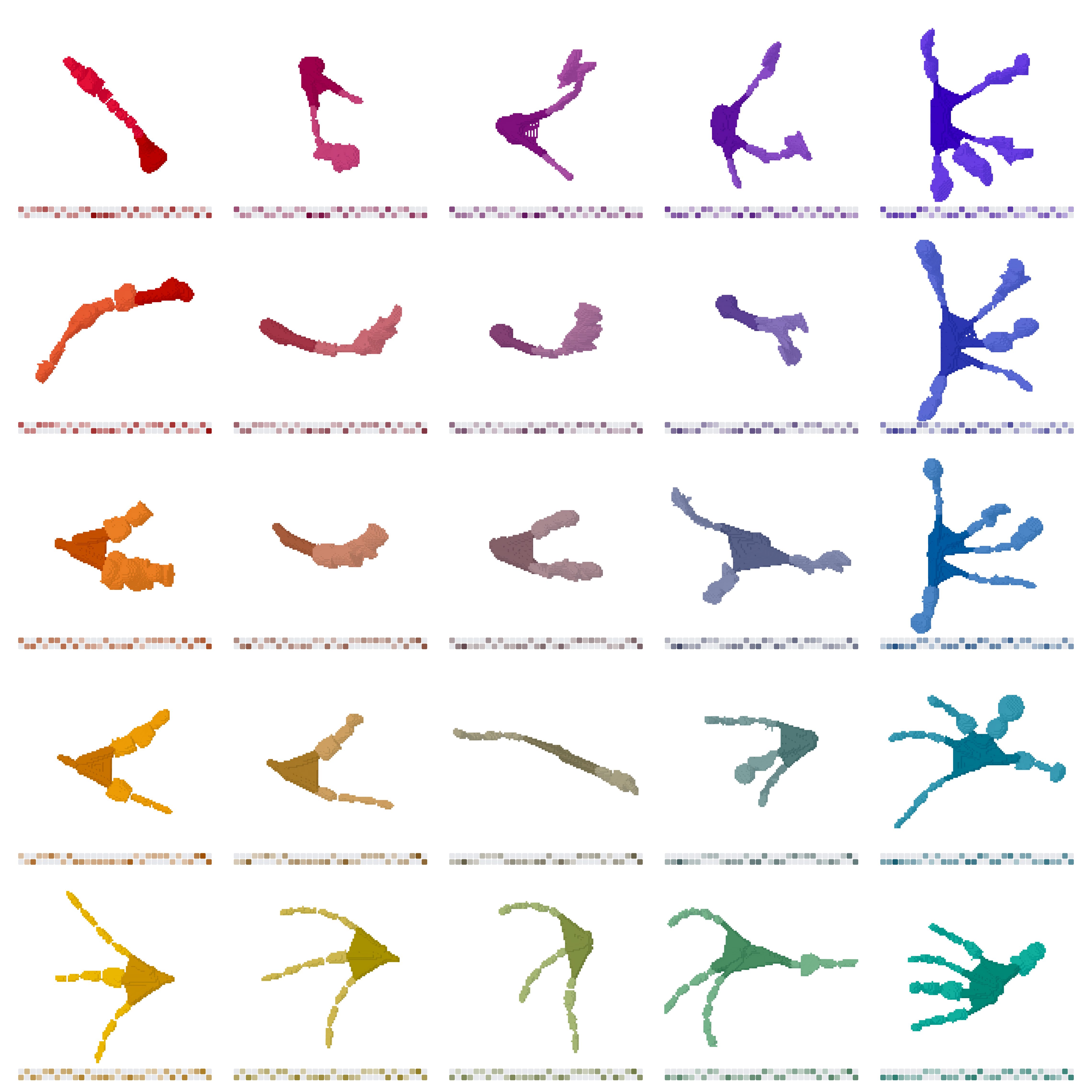}
  \vspace{-18pt}
  \caption{\textbf{Genetic interpolation between four evolved manipulators.}
  Spherical linear interpolation (slerp) between 
  a retriever (top left), 
  a rotator (top right), 
  a picker (bottom right), 
  and a hanger (bottom left).
  The latent genotype, $z_i \in \mathbb{S}^{31}$, below each decoded design is visualized as a double barcode.
  The upper and lower rows of the barcode show the positive and negative component of each latent dimension, respectively, with darker colors indicating greater magnitude.
    }
    \vspace{-1em}
  \label{fig:latent-interp}
\end{figure*}

\begin{figure*}[!t]
  \centering
  \includegraphics[width=\columnwidth]{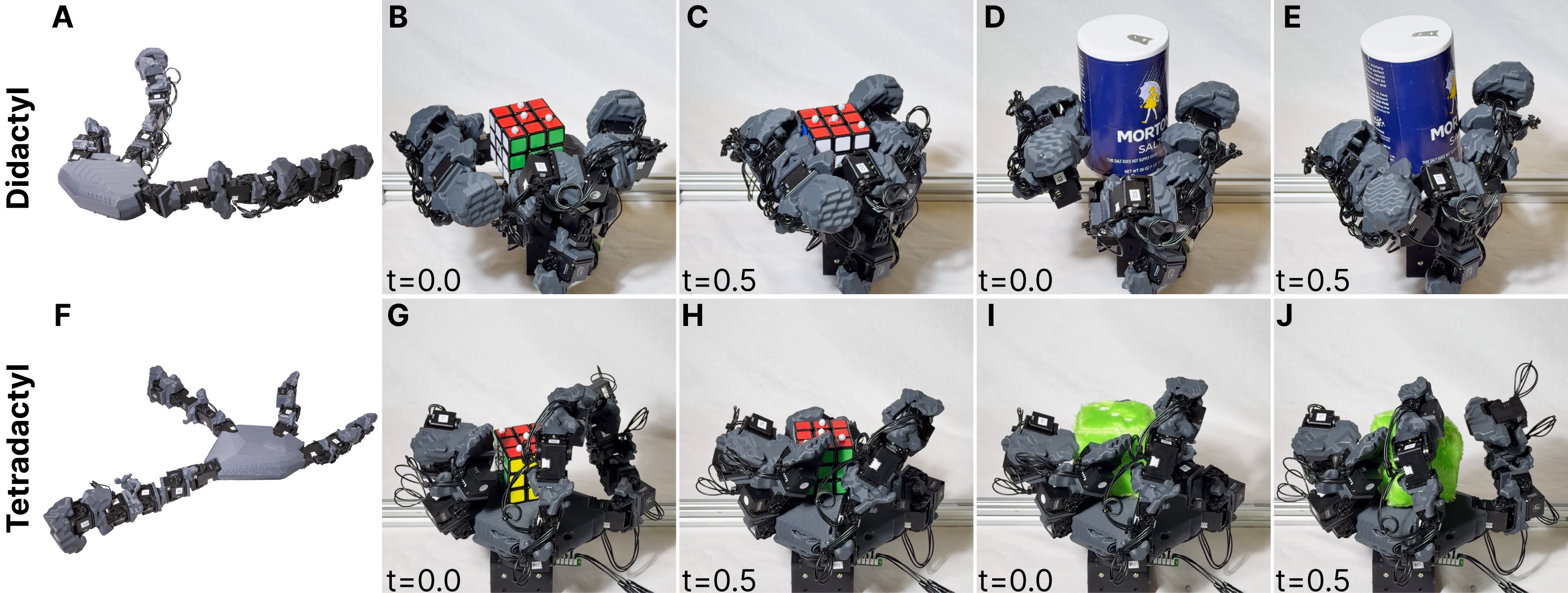}
  \vspace{-18pt} 
  \caption{\textbf{Two evolved rotators.}
  In our main experiment, robots were evolved to rotate diverse objects. 
  The best was essentially didactyl (\textbf{A-E}).
  It possessed two functional appendages and a third, very short vestigial appendage that did not contribute to object rotation.
  Another highly fit evolved design which we built 
  was tetradactyl (\textbf{F-J});
  all four of its appendages were utilized.
  }
  \vspace{-1em}
  \label{fig:robot_gallery_rotation}
\end{figure*}

\section{Methods}
\label{sec:methods} 

In this section we describe
how designs were 
indexed (Sect.~\ref{sec:morphology}),
encoded and decoded (Sect.~\ref{sec:methods-latent-genome}),
translated into to a realizable form (Sect.~\ref{sec:translating}),
trained (Sect.~\ref{sec:RL}),
evolved (Sect.~\ref{sec:evolution}),
manufactured
and tested in reality~(Sect.~\ref{sec:reality}).

\subsection{Morphospace}
\label{sec:morphology}


Designs were parameterized by
a root (Fig.~\ref{fig:development}I) and 
up to five appendages (kinematic chains) that may grow outward from any point along the edge of the root (Fig.~\ref{fig:development}L).
Each design contained up to 25 joints.
Its root was parameterized by 39 variables (Fig.~\ref{fig:design_space-root}) and each segments along an appendage was parameterized by 12 variables (Fig.~\ref{fig:design_space-appendage}).
A compiler converts these variables into a body plan, voxelized on a 128-by-128-by-128 grid with two channels.
The first channel indicates which voxels belong to the root; non-root voxels were set to -1.
The second channel indicates topological distance from the root using relative coordinates, where  voxels within proximate and distal segments were indexed as 0 and 1, respectively. 
Each intermediate segment, $k$, along an appendage with $K$ segments was indexed as $k/(K-1)$.
Non-appendage voxels in the second were set to $-1$. 
Finally, 
a kinematic graph stores the relative location and hinge axis of each joint.

\subsection{Latent Space (The Genome)}
\label{sec:methods-latent-genome}


We used contrastive learning \citep{pmlr-v119-chen20j} to create a latent genome in which
similar morphologies have similar genes
and dissimilar morphologies have genes that are farther away in the latent space.
The voxel grid was encoded by a 3D convolutional network \citep{maturana2015voxnet}
and
the kinematic tree was encoded by a graph network \citep{ying2018hierarchical}.
The two inputs streams were combined into a single latent vector $z$ and normalized onto the surface of a hypersphere
\citep{pmlr-v119-wang20k}.
More information about the encoder can be found in Appx.~\ref{app:encoder}.

The decoder maps a latent genotype to a sequence of design tokens (Fig.~\ref{fig:development}H)
that specify the morphology.
Continuous morphological parameters were quantized into multiple tokens.
An autoregressive transformer \citep{vaswani2017attention} conditioned on the latent genotype vector 
predicts the sequence 
one token at a time (Fig.~\ref{fig:development}B).
During body generation, 
a grammar masks erroneous tokens such that every body is decoded from a complete and valid set of morphological parameters.
More information about the decoder can be found in Appx.~\ref{app:decoder}.

Two million example body plans were generated
and each was converted into a ground-truth token sequence, $s$. 
(See Appx.~\ref{app:synthetic-dataset} for details.)
This synthetic data was used to optimize the encoder and decoder simultaneously with the loss function
$
    \mathcal{L} = \mathcal{L}_{\mathrm{con}} + \lambda \, \mathcal{L}_{\mathrm{rec}}.
$
The contrastive term, 
$\mathcal{L}_{\mathrm{con}}$, 
uses NT-Xent loss \citep{pmlr-v119-chen20j} to update the encoder.
For each example design, $z$, a morphologically similar counterpart, $z'$, was generated through
small perturbations and transformations 
to the original body plan,
rescaling, reorienting, repositioning, or slightly deforming it on the voxel grid 
or removing distal segments.
These ``positive pairs'' ($z$ and $z'$) are brought closer together and farther away from other examples (``negative pairs'').
The reconstruction term,
$\mathcal{L}_{\mathrm{rec}}$, 
uses next-token cross-entropy 
to optimize the decoder.
See Appx.~\ref{app:latent-training} for more details.


\subsection{Translating Designs into Robots}
\label{sec:translating}

After a design was decoded 
from latent space to voxel space,
it was automatically converted into a manufacturable form \citep{guo2026creating}.
A printable mesh was extracted 
\citep{lorensen1987marching}
for each body part  
and servos 
were placed at every joint. 
If all of the motors did not fit into the body, the design was rejected.
Body parts were carved out slightly to accept the hardware, with
cavities for each motor and its mounting bracket, and channels for screws and cables.
To prevent self-collision, each joint was swept through its range of motion and any material of either segment that would collide was removed.
This sweep stopped before a segment lost too much of its volume or was broken apart into disconnected pieces, 
and this safe range became the joint limit.
Finally, the motors and their brackets were merged into the body model together with their masses.
The resulting, assembly-ready robot was exported as a URDF file for simulation
and as set of meshes for printing.
%

\subsection{Learning to Manipulate Diverse Objects}
\label{sec:RL}

Robot were trained in Isaac Lab \citep{mittal2025isaac}.
Before policy training, 
a neural pose was found in simulation by optimizing joint angles such that enough segments get close to the object without self-collisions.
If and when a valid neutral pose could not be found, the design was discarded. 
A control policy was then trained from scratch for each design with proximal policy optimization (PPO; \cite{schulman2017proximal}) 
across thousands of parallel environments.
The policy observes joint angles, previous commands, and the state of the object, 
and outputs a target position for every joint.
After training, the optimized policy was replayed without exploration noise for a fixed window,
and fitness was taken to be average reward achieved across environments.
See Appx.~\ref{app:tasks-details-entire-section} for details.

\subsection{Evolving Novel Robots}
\label{sec:evolution}

Following \cite{li2025generating},
robots were designed by covariance matrix adaptation evolutionary strategies (CMA-ES; 
\cite{hansen2001completely}).
Briefly, genotypes were sampled from a normal distribution in the latent space.
During evolution, the distribution is pulled toward sampled latents with the highest fitness, stretched along the most promising dimensions of covariation, and compressed along others.
Depending on how far the distribution mean is moving across the latent space, the step size is either increased or decreased, dynamically.
For more details see \cite{hansen2016cma}.
Hyperparameters can be found in Table~\ref{tab:cmaes_hyperparameters}.

\subsection{Reality Check}
\label{sec:reality}

Winning designs were fabricated directly from their automatically generated blueprints.
Body parts were 3D-printed using PLA (polylactic acid) filament
and the Dynamixel servos were inserted and attached by screws into heat inserts.
(A complete bill of materials can be found in Table.~\ref{tab:bom}.)
During evolution the policy observes privileged information, i.e.~the state of the object. 
The policy of the winning design was distilled into a proprioceptive controller that observes only the joint positions and previous commands,
and was finetuned under domain randomization (Table~\ref{tab:domain-randomization}).
The finetuned policy was deployed zero-shot on the physical robot at the same control rate as used during training in simulation (30~Hz).  
See Appx.~\ref{app:tasks-details-entire-section} for details.

\begin{figure*}[!t]
  \centering 
  \includegraphics[width=\columnwidth]{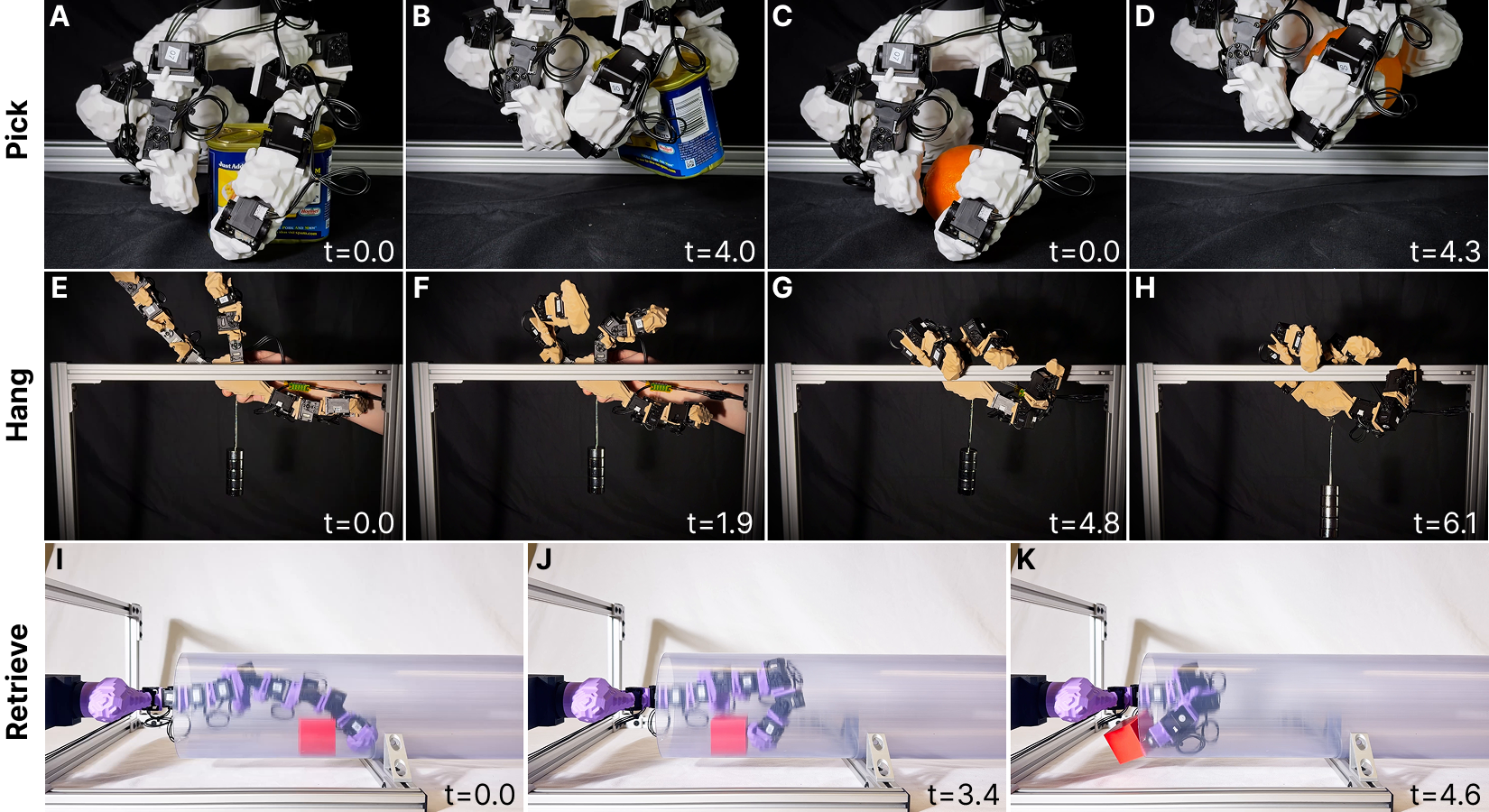}
  \vspace{-18pt}
  \caption{\textbf{An evolved picker, hanger, and retriever.}
  We evolved, trained, and realized robots that 
  pick objects off a surface~(\textbf{A-D}), 
  hang from a horizontal bar with a weight suspended below them~(\textbf{E-H}),
  and 
  retrieve objects from within a narrow pipe~(\textbf{I-K}).
  }
  \vspace{-1em}
  \label{fig:robot_gallery_pick_hang_retrieve}
\end{figure*}

\section{Results}
\label{sec:results}

We being this section by examining the latent genome (Sect.~\ref{sec:results-genomics}).
Next we consider six tasks:
rotating (Sect.~\ref{sec:rotate}),
retrieving (Sect.~\ref{sec:retrieve}),
flipping (Sect.~\ref{sec:flip}),
and picking (Sect.~\ref{sec:pick}) objects,
holding a tool (hammer; Sect.~\ref{sec:hold}),
and hanging from the environment (a bar; Sect.~\ref{sec:hang}).
We evolved specialists for each task and generalists capable of multiple tasks (Sect.~\ref{sec:generalist}).
Finally, we conduct an ablation study to determine how different genetic encodings may help or hinder evolution (Sect.~\ref{sec:ablation}). 

Although it is difficult to decompose animal behavior into discrete ``tasks'', 
the capacity to retrieve, pick, and hold objects (e.g.~food) is clearly important for survival.
Likewise, gripping and hanging rather than falling from objects (e.g.~a tree branch) can be a matter of life or death.
More sophisticated behaviors require reorienting (rotating, flipping) the object.
It can be equally important---and very interesting---for robots to evolve anatomies that support these vital behaviors.

\subsection{Genomics}
\label{sec:results-genomics}

In order to evaluate the smoothness and coherence of the latent genome, 
we generated a test set of 2,000 unique body plans that were not used for training and sequenced their latent genotypes.
For each design we created an augmented
%
%
copy and measured how often the copy and original were nearest genetic neighbors.
We refer to this as ``recall'' (in Table~\ref{tab:latent-ablation}) and found that indeed all original-modified pairs were closer together than to any other design, a recall of 100\%.
We also measured, on a separate test set of 4,500 designs spanning 17 topologies, 
how often the kinematic structure of each design matched the most common structure among its 10 nearest genetic neighbors, a statistic we term ``topology 10-NN accuracy'', and found it to be 92\%.
Next we created amputated copies of the test designs with up to four distal segments removed
and
measured the
correlation (Spearman's)
between how many segments were amputated and 
the genetic distance 
from the original
(edit-distance correlation; $\rho=0.97$). 
Finally, we encoded and decoded designs and measured how often the decoded design had the expected number of appendages (appendage accuracy; 100\%) and joints (joint accuracy; 96\%), as well as how often the decoded design stayed within the voxel grid with nonoverlapping appendages (validity; 82\%). See Table~\ref{tab:latent-ablation} for complete analysis.

\subsection{Rotate}
\label{sec:rotate}

We first considered ``in hand'' rotation, a standard manipulation benchmark \citep{handa2023dextreme,yin2023rotating,shaw2023leaphand,fay2026house}.
The robot was mounted to the environment with its root facing upward to accept an object.
The objective is to vertically rotate the object placed on top of the root, i.e.~in its ``palm'' (Fig.~\ref{fig:evolution}A-D).
Fitness was the net rotation rate of the object over a 30 second evaluation period.
Five independent evolutionary trials were conducted and 
compared against 
an anthropomorphic baseline \citep{shaw2023leaphand} 
as well as
\cite{fay2026house}, 
who optimized manipulators for the same task and
which, until now, represented the state-of-the-art (Fig.~\ref{fig:fitness-curves}).
More details about the task environment and reward function used for policy training can be found in Appx.~\ref{app:task-details-rotate}.

We realized two evolved designs (Fig.~\ref{fig:robot_gallery_rotation}).
The policy learned during evolution was
distilled 
so as to no longer rely on privileged (object state) information
and finetuned under domain randomization (Table~\ref{tab:domain-randomization}).
After finetuning, the evolved didactyl robot rotated objects at 3.66 rad/s in simulation,
compared to 
2.09 rad/s on the anthropomorphic hand \citep{shaw2023leaphand} 
and
3.3 rad/s reported by \cite{fay2026house} using privileged information.
Deployed zero-shot in the real world, the didactyl robot rotated a diverse set of previously unseen objects (weighing from 16 to 687~g) 
at 1.39 rad/s on average,
almost twice as fast as the best design reported by \cite{fay2026house}, which achieved 0.74 rad/s on average.
The evolved tetradactyl robot rotated cubes at 1.78 rad/s on average but failed to rotate the other test objects.
After the same training process,
the anthropomorphic hand \citep{shaw2023leaphand} failed to rotate any of the previously unseen test objects.
Even with a specialized policy trained to rotate a cube,
the hand was no better than the evolved designs at rotating cubes.
See Table~\ref{tab:real_world_rotation} for details.



\begin{figure*}[!t]
  \centering
  \includegraphics[width=\columnwidth]{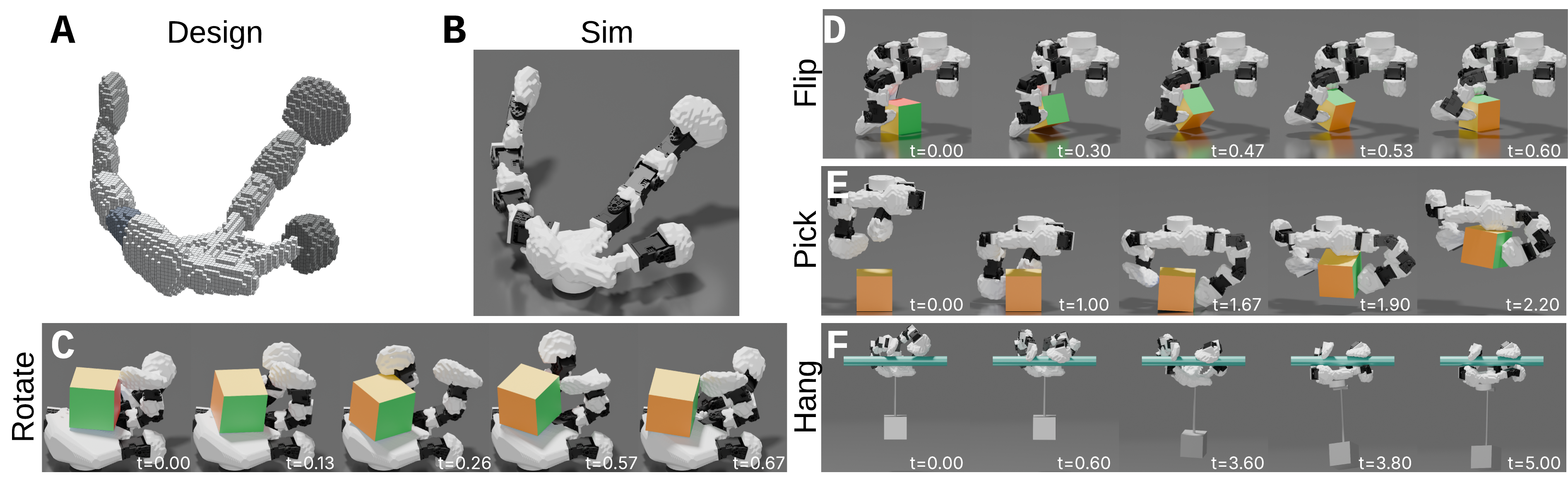}
  \vspace{-18pt} 
  \caption{\textbf{An evolved multitasker.}
  We also evolved a generalist to carry out several tasks. 
  The winning design (\textbf{A,B}) could 
  rotate (\textbf{C}), 
  flip (\textbf{D}), 
  pick up (\textbf{E}), 
  and hang (\textbf{F}) from 
  objects in simulation.
  }
  \vspace{-1em}
  \label{fig:robot_gallery_generalist}
\end{figure*}


\subsection{Pick}
\label{sec:pick}


Robots were mounted 
above a tabletop
with their root facing downward  and evolved to pick up objects (Fig.~\ref{fig:evolution}O-R).
Highly fit robots pick the object and keep it from falling in the under
external perturbations (see Appx.~\ref{app:task-details-pick}
for more details). 
We conducted a single evolutionary trial; the winning design picked up and held onto objects 99\% of the time.
We realized this design (Figs.~\ref{fig:evolution}Q and \ref{fig:robot_gallery_pick_hang_retrieve}A-D),
and deployed its policy zero-shot after the same distillation and finetuning procedure described above.
Across 24 previously unseen objects, 
it succeeded in eight out of ten attempts on average,
and picked up every object at least once except the heaviest, a 687\,g salt container
(Table~\ref{tab:real_world_grasping}).

\subsection{Hang}
\label{sec:hang}

Robots were positioned near a horizontal bar and evolved to wrap around it such that, when released, the body hangs from the bar with a 500\,g weight suspended below it (Fig.~\ref{fig:evolution}S-V).
%
Fitness was based on how long the robot stayed on the bar, discounted by the motor torque expended during an evaluation period of 20 sec
(see Appx.~\ref{app:task-details-hang}
for details). 
The winning design held on for the entire evaluation period in 99.9\% of training episodes,
using half the torque of the best designs found earlier in evolution.
We realized this design (Fig.~\ref{fig:evolution}S); deployed zero-shot with a finetuned policy, it gripped and hung from a 
bar under a 500\,g load~(Fig.~\ref{fig:robot_gallery_pick_hang_retrieve}E-H).

\subsection{Retrieve}
\label{sec:retrieve}

Robots were mounted at the opening of a narrow horizontal pipe containing a small cube~(Fig.~\ref{fig:evolution}E-H). 
Robots were evolved to reach into the pipe and pull the cube out (see Appx.~\ref{app:task-details-retrieve}
for details).
The winning design, which was successful in every training episode, was realized, finetuned, and deployed zero-shot as described above. 
It was equally successful in reality (Fig.~\ref{fig:robot_gallery_pick_hang_retrieve}I-K).

\subsection{Flip}
\label{sec:flip}

Robots were mounted above a table and evolved to flip a cube over, again and again, in a randomly chosen direction, as quickly as possible.
Each flip moves the cube one edge length along the table,
so the mounted robot must also pull the cube back within reach to keep flipping it.
Fitness was the number of full rotations of the cube over a 60 sec evaluation period (see Appx.~\ref{app:task-details-flip}
for details).
The winning design (Fig.~\ref{fig:evolution}I-K) achieved 7.75 full rotations in 60 sec.

\subsection{Hold}
\label{sec:hold}

The manipulator was mounted on the end of an arm and evolved to hold a hammer by its handle 
as arm swings the hammer up and down 
(Fig.~\ref{fig:evolution}N).
Fitness was how little the hammer moved relative to its mounted root during arm swinging and under external perturbations that pulled the hammer in random directions  (see Appx.~\ref{app:task-details-hold}
for details).
The winning design (Fig.~\ref{fig:evolution}L) dropped the hammer in just 3\% of episodes.
It was the only winning design with a pentadactyl morphology that evolved
across the six task environments we tested.


\begin{figure*}[!t]
  \centering
  \includegraphics[width=\columnwidth]{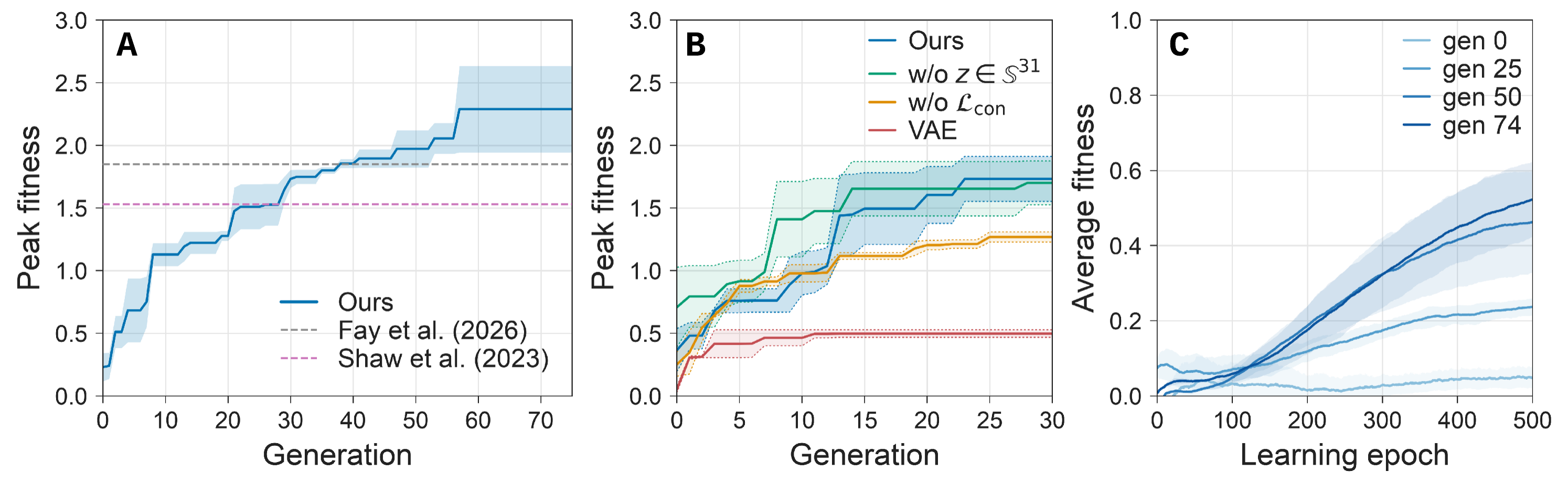}
  \vspace{-18pt} 
  \caption{\textbf{The evolution of rotators}
  (fitness measured during evolution; results after finetuning are in Sect.~\ref{sec:rotate}). Our approach consistently yields novel robots that rotate objects faster than 
  those produced by related prior work, where our full model was trained with the 2 million sample dataset
  (\textbf{A})
  and
  the proposed genetic representation increased evolvability beyond what could be achieved with a VAE, both trained in the smaller 200,000 sample dataset (\textbf{B}).
  In the full setting, morphologies evolved that made learning much easier (\textbf{C}).
  Shaded regions indicate standard error of the mean.
  Here results are reported before policy distillation and finetuning.
  }
  \vspace{-1em}
  \label{fig:fitness-curves}
\end{figure*}


\subsection{Multitasking}
\label{sec:generalist}

Finally, we evolved manipulators to perform five of the six tasks described above (all but retrieve).
Fitness was based on performance across all five tasks
(see Appx.~\ref{app:task-details-multitask} for details).
Early in evolution, the best design in the population had mastered object rotation and flipping and hanging from the bar.
Later, the best design evolved the ability to pick and hold objects but as it did it got worse at rotating them.
The final, winning design  (Fig.~\ref{fig:robot_gallery_generalist}) 
was almost as good at flipping, picking, and hanging 
as the specialists evolved for those tasks alone,
but it rotated objects slowly and dropped the hammer.

\subsection{Ablations}
\label{sec:ablation}

We compared our latent embedding against three alternatives (Table~\ref{tab:latent-ablation}).
The first two were otherwise equivalent but 
lacked the spherical projection (w/o $z \in \mathbb{S}^{31}$) 
or contrastive loss (w/o $\mathcal{L}_{\mathrm{con}}$).
The third alternative was a VAE \citep{hu2023glso,li2025generating,mitchell2025genomic,yu2026reconfigurable}.
All were trained (ours was retrained) with a smaller set of 200,000 example body plans
and share the same encoder and decoder architecture.
Three evolutionary trials were conducted for the task of object rotation (Fig.~\ref{fig:fitness-curves}B).
Without contrastive learning,
winning designs were significantly worse.
The VAE suffered posterior collapse, 
rarely reconstructing the correct number of appendages and joints,
and so evolution quickly fell into a local optimum.
See Appx.~\ref{app:ablation_analysis} for complete analysis.

%
%

\section{Discussion}
\label{sec:discussion}

In this paper, we evolved morphologically complex robots to 
flexibly handle diverse objects across a wide array of tasks (Fig.~\ref{fig:evolution}).
The resulting robots were shown to be more dexterous than
those generated by otherwise equivalent state-of-the-art baselines (Fig.~\ref{fig:fitness-curves})
which 
operate within a more restrictive design space \citep{fay2026house} 
or 
assume an anthropomoric layout \citep{shaw2023leaphand}.
We used contrastive learning to map the combinatorial space of possible designs onto 
a spherical latent genome in which
similar body plans were genetically closer than dissimilar bodies (Fig.~\ref{fig:latent-ablation}) and an autoregressive
decoder to retain finegrain morphological features and ensure validity (Fig.~\ref{fig:development}).
This genetic representation significantly increased the evolvability of robots
beyond the best known
alternative~(Fig.~\ref{fig:fitness-curves}), 
which relied 
on a simple but effective VAE framework of
reconstruction and regularization
\citep{mitchell2025genomic}.
Although it sounds plausible in retrospect that contrastive learning would yield a latent space that aids morphological evolution, 
this result was quite surprising; 
distinct designs could have been inadvertently clustered on disjoint genetic islands, obstructing the evolution of novel forms.
On the contrary, we found contrastive loss to be essential; without it, manipulators were significantly less evolvable~(Fig.~\ref{fig:fitness-curves}). 
\cite{kirchoff2024salsa} used a similar approach to design drugs based on their molecular graphs
but they did not test if the sphere projection mattered. 
Anecdotally, this projection slightly increased evolvability on average,
though this comparison was not statistically significant
and the dis/advantages for evolving robots remain unclear.




There were, of course, several important limitations of these experiments.
The design space was limited to a one material (PLA), 
one mode of actuation (in-joint servos), 
and a single sensor modality (proprioception);
evolution and learning were conducted entirely in simulation without feedback from the real environment;
the desired behaviors were cognitively simple and could be executed without prospection or planning;
and the robots themselves were sessile, they could only handle nearby objects---everything else was out of reach.
Future work will study the evolution of more thoroughly sensorized, multi-material agents that move themselves from place to place, interacting with objects along the way and learning from this real world experience.

It is also important to note the necessity of simulation; without it, the evolution remains prohibitively expensive for all but the simplest robots \citep{brodbeck2015morphological}.
An interesting open question is therefore whether and how relevant morphology-conditioned dynamics
that are 
mismodeled by or missing from 
the simulator
may be learned on the fly 
with each deployed robot \citep{o2022neural,matthews2026zero}.

Despite these and many other limitations,
the dexterous robots in this paper were not limited by human preconceptions about what a manipulator should look like;
rather,
they were free (more or less) to evolve whatever structural and architectural features they needed to solve the task at hand.


\section*{Acknowledgments}

This research was supported by
NSF awards 2331581 and 2440412,
and TWCF award 20650.

\bibliography{main}
\bibliographystyle{iclr2027_conference}

\clearpage
\appendix

\etocdepthtag.toc{mtappendix}
\etocsettagdepth{mtchapter}{none}
\etocsettagdepth{mtappendix}{subsection}
\etocsettocstyle{\section*{Appendix}}{}
\tableofcontents

\clearpage
\section{Design space}
\label{app:design_space}

\begin{figure}[!ht]
  \centering
  \includegraphics[width=\columnwidth]{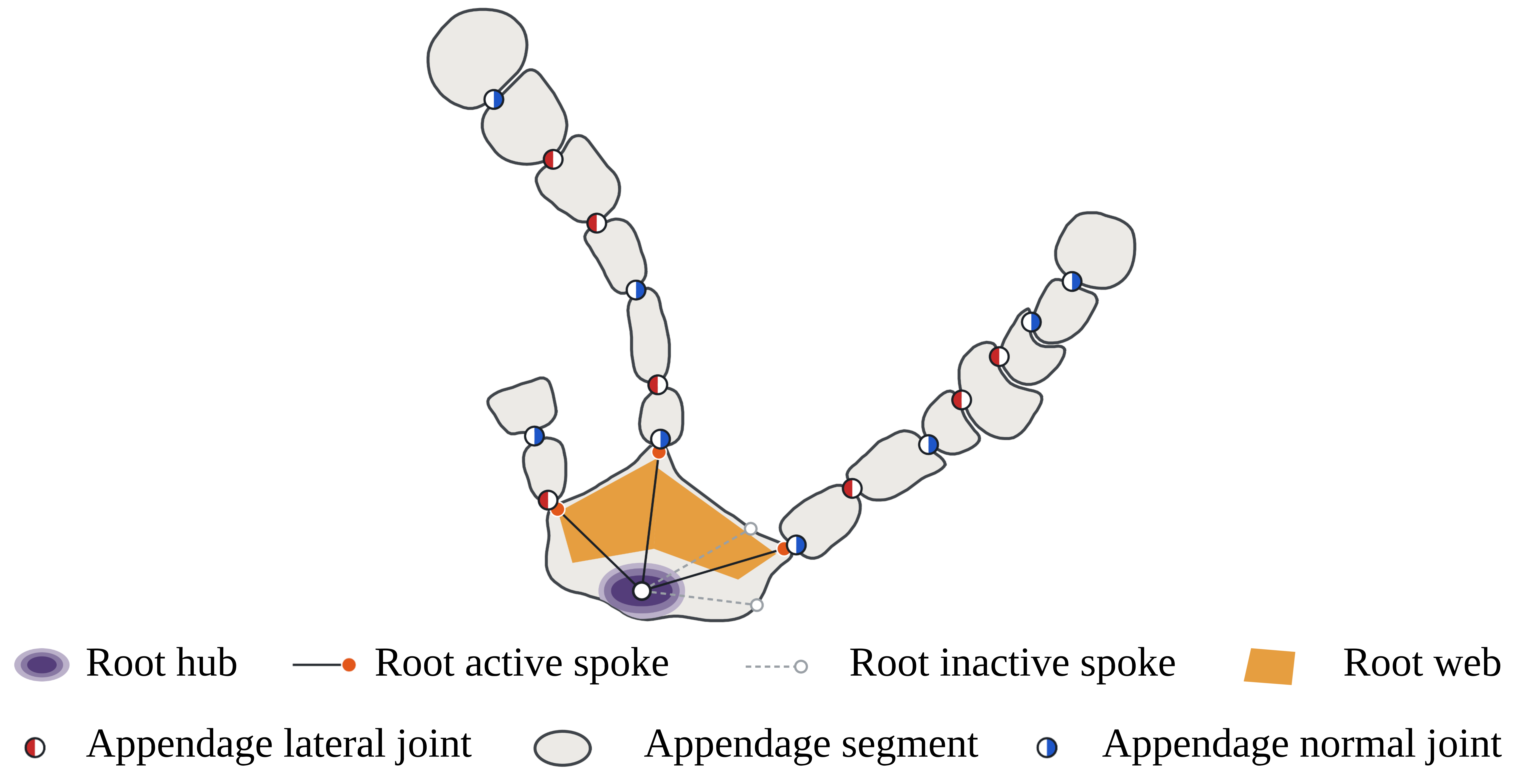}
  \caption{\textbf{One design.} The root is the central body: the hub at its
centre and five spokes around it. Three spokes are active and so built, each ending at the tip from which an
appendage grows; two are inactive and not built.
 A web fills the space
between neighbouring built spokes. Each appendage is a chain of segments, and
each joint between two segments is marked by the hinge axis it takes, lateral
or normal. Segments are drawn slightly apart so that every joint shows. The
parameters of the root and of an appendage are shown one by one in
Figs.~\ref{fig:design_space-root} and~\ref{fig:design_space-appendage}.}
  \label{fig:design_space-overview}
\end{figure}

A design is made of one root and up to five appendages
(Fig.~\ref{fig:design_space-overview}). The root is the central body. It has
three components: the hub, a flat elliptical disc at its centre; five spokes
that radiate from the hub, each either built or not; and the webs, which close
the gap between neighbouring built spokes. Its 39 parameters
(Table~\ref{tab:design_space_ranges}) set the size and shape of the hub and
the direction, length, tilt and thickness of each spoke, so that the root can
range from a thin flat frame to a thick cupped body, with its appendages
starting at any angle around it (Appx.~\ref{app:design_space_root}).

One appendage grows from the tip of each built spoke. An appendage is a chain
of segments that never branches, each segment joined to the one before it by
a joint. Each segment, with its joint, has 12 parameters, which set how its joint turns and
where it points, and the length, cross-section and surface of the segment
itself (Appx.~\ref{app:design_space_link}). A design has at most 25 segments,
and so at most 25 joints.

A design is turned into a voxelized body in three steps. First, the parameters
place a skeleton: the hub, each spoke from its foot to its tip, and each
chain of segments, every segment starting where the one before it ends.
Second, every part of the skeleton is given a solid shape, and the shapes that
make up one part are blended smoothly into each other. Third, the body is
voxelized on a $128^3$ grid, with every voxel labelled by the part it belongs
to.

\subsection{The root}
\label{app:design_space_root}

\begin{figure}[!ht]
  \centering
  \includegraphics[width=\columnwidth]{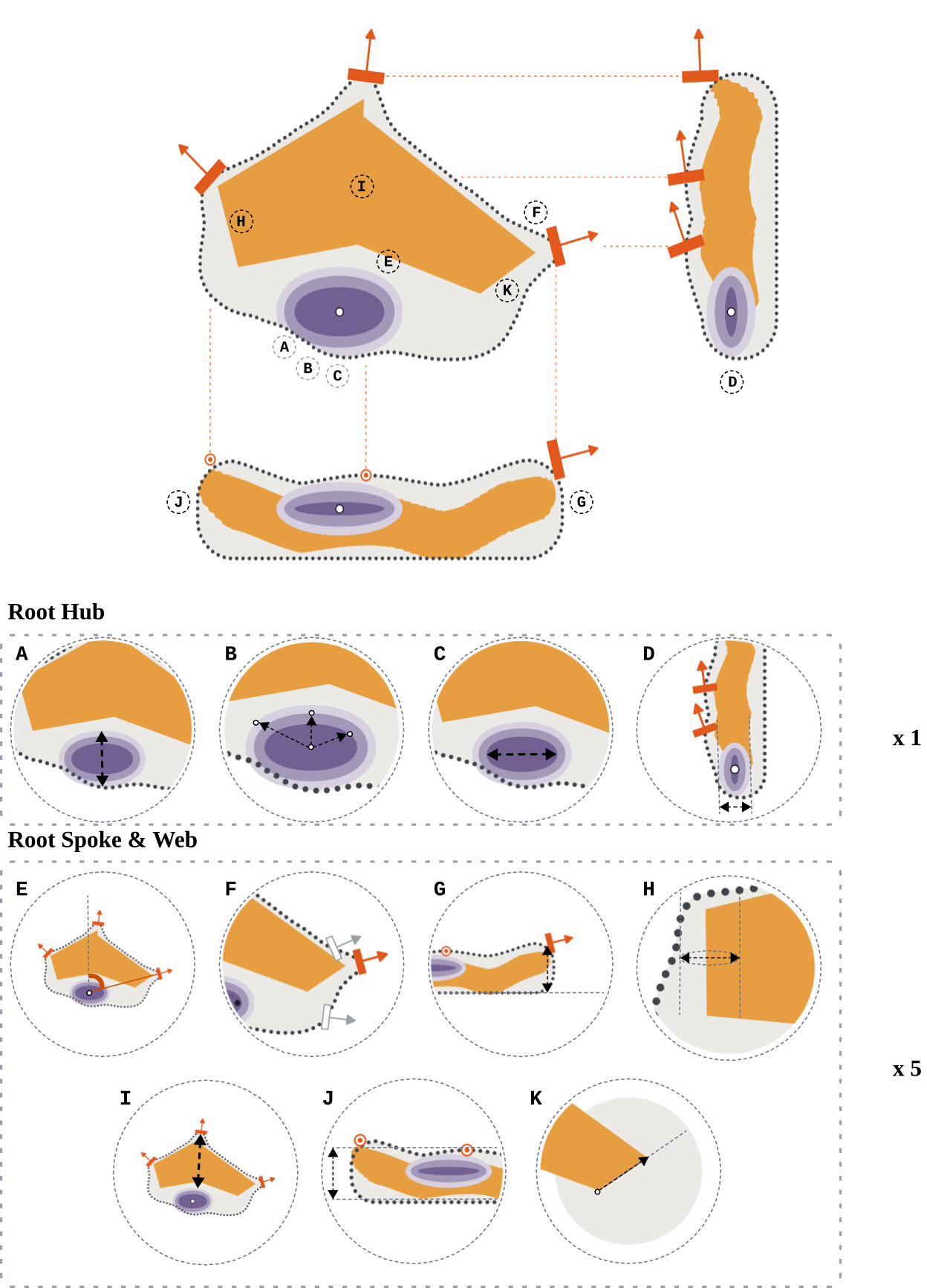}
  \caption{\textbf{The root.} Top: one root from the design space, seen from
above (left), from the end (right) and from the side (bottom). The purple
ellipse is the hub. An orange bar with an arrow marks the tip of each spoke
that is built. \textbf{A--K}: one parameter each, cut and enlarged from the
views above; the small circle on the views marks where each was cut.
Root hub: A \textbf{Root hub aspect}, B \textbf{Root hub ring fraction},
C \textbf{Root hub radius}, D \textbf{Root hub infill}. Root spoke and web:
E \textbf{Root spoke azimuth}, F \textbf{Root spoke activation}, G \textbf{Root spoke
elevation}, H \textbf{Root spoke radius}, I \textbf{Root spoke length},
J \textbf{Root web radius}, K \textbf{Root web reach}.}
  \label{fig:design_space-root}
\end{figure}

The root has three parts: the hub at its centre, the five spokes around it,
and the webs between neighbouring built spokes.
The letter after each name is its panel in Fig.~\ref{fig:design_space-root}.

The hub has four parameters.
\begin{list}{}{\setlength{\labelwidth}{4.7cm}\setlength{\labelsep}{0.2cm}\setlength{\leftmargin}{4.9cm}\setlength{\itemsep}{2pt}\setlength{\parsep}{0pt}\setlength{\topsep}{4pt}}
  \item[{\makebox[4.0cm][l]{\textbf{Root hub aspect}}\makebox[0.7cm][l]{(A)}}]
        stretches the hub from a disc into an oval: round at 1, and three times as long as wide at 3.
  \item[{\makebox[4.0cm][l]{\textbf{Root hub ring fraction}}\makebox[0.7cm][l]{(B)}}]
        sets where the feet of the spokes sit on the hub, from halfway out (0.5) to the rim (1.0).
  \item[{\makebox[4.0cm][l]{\textbf{Root hub radius}}\makebox[0.7cm][l]{(C)}}]
        sets the size of the hub, and with it how far apart the feet of the spokes are.
  \item[{\makebox[4.0cm][l]{\textbf{Root hub infill}}\makebox[0.7cm][l]{(D)}}]
        sets how much material the hub carries, from a thin frame between the feet of the spokes to a thick solid body.
\end{list}

Each of the five spokes, with its web, has seven parameters.
\begin{list}{}{\setlength{\labelwidth}{4.7cm}\setlength{\labelsep}{0.2cm}\setlength{\leftmargin}{4.9cm}\setlength{\itemsep}{2pt}\setlength{\parsep}{0pt}\setlength{\topsep}{4pt}}
  \item[{\makebox[4.0cm][l]{\textbf{Root spoke azimuth}}\makebox[0.7cm][l]{(E)}}]
        sets where around the hub the spoke comes out, and so where its appendage starts.
        Spokes with similar azimuths sit side by side; spokes half a turn apart face each other across the hub.
  \item[{\makebox[4.0cm][l]{\textbf{Root spoke activation}}\makebox[0.7cm][l]{(F)}}]
        decides whether the spoke is built: it is built if its activation is greater than 0,
        and such a spoke is called active; otherwise it is inactive and not built.
        If no activation is greater than 0, the spoke with the largest activation is built,
        so a design always has at least one appendage.

  \item[{\makebox[4.0cm][l]{\textbf{Root spoke elevation}}\makebox[0.7cm][l]{(G)}}]
        tilts the spoke out of the plane of the hub, so that the appendage starts above or below the hub.
        Tilting all spokes the same way cups the root.
  \item[{\makebox[4.0cm][l]{\textbf{Root spoke radius}}\makebox[0.7cm][l]{(H)}}]
        sets how thick the spoke is. A spoke is thickest at its foot and narrows toward its tip.
  \item[{\makebox[4.0cm][l]{\textbf{Root spoke length}}\makebox[0.7cm][l]{(I)}}]
        sets how far from the centre of the hub the tip of the spoke lies, and so how far out the appendage starts.
        The length of a spoke, from foot to tip, is kept between 3.5 and 16\,cm.
  \item[{\makebox[4.0cm][l]{\textbf{Root web radius}}\makebox[0.7cm][l]{(J)}}]
        sets how thick the web is. A thin web leaves the spokes distinct; a thick web merges them into one plate.
  \item[{\makebox[4.0cm][l]{\textbf{Root web reach}}\makebox[0.7cm][l]{(K)}}]
        sets how far along the spoke the web extends toward the tip.
        The web between two spokes goes only as far as the smaller of their two values.
\end{list}
A spoke that is not built still places its foot on the hub,
so its azimuth and elevation, and where its foot sits, still shape the hub,
while its length, radius and web reach have no effect.

\begin{table}[!htb]
  \centering
  \small
  \setlength{\tabcolsep}{4pt}
  \caption{Design-space parameters. The root is described by 39 values and
  each segment of an appendage, with its joint, by 12. Rows follow the panels of
  Figs.~\ref{fig:design_space-root} and~\ref{fig:design_space-appendage}.
  Bins is the number of bins used to write the parameter as a token
  (Appx.~\ref{app:decoder}).}
  \label{tab:design_space_ranges}
  \begin{tabular}{|l|l|c||l|l|c|}
    \hline
    \textbf{Parameter} & \textbf{Range} & \textbf{Bins} & \textbf{Parameter} & \textbf{Range} & \textbf{Bins} \\
    \hline
    \multicolumn{3}{|l||}{\emph{Root hub}} & \multicolumn{3}{l|}{\emph{Appendage joint}} \\
    \hline
    root hub aspect          & 0.8 to 3.0              & 12  & appendage joint yaw bend        & $-0.24$ to $0.24$ rad & 32 \\
    root hub ring fraction   & 0.5 to 1.0              & 12  & appendage joint hinge axis      & lateral or normal     & 2  \\
    root hub radius          & 0.02 to 0.05 m          & 12  & appendage joint pitch bend      & $-0.26$ to $0.26$ rad & 32 \\
    root hub infill          & 1.0 to 4.0              & 8   & appendage joint length          & 0.03 to 0.06 m        & 32 \\
    \hline
    \multicolumn{3}{|l||}{\emph{Root spoke and web}} & \multicolumn{3}{l|}{\emph{Appendage segment}} \\
    \hline
    root spoke azimuth       & $-\pi$ to $\pi$ rad     & 128 & appendage segment asymmetry     & 0.0 to 2.0            & 8  \\
    root spoke activation    & built if $> 0$          & 2   & appendage segment gain  & 0.55 to 1.20          & 8  \\
    root spoke elevation     & $-1$ to $1$             & 64  & appendage segment lobe          & 0.40 to 1.10          & 8  \\
    root spoke radius        & 0.01 to 0.03 m          & 8   & appendage segment phase & $-\pi$ to $\pi$ rad   & 16 \\
    root spoke length        & 0.06 to 0.10 m          & 32  & appendage segment width         & 0.01 to 0.04 m        & 12 \\
    root web radius          & 0.6 to 1.2              & 8   & appendage segment depth         & 0.01 to 0.04 m        & 24 \\
    root web reach           & 0.7 to 1.0              & 8   & appendage segment taper         & 0.10 to 1.5           & 12 \\
                             &                         &     & appendage segment twist         & $-1.20$ to $1.20$ rad & 8  \\
    \hline
  \end{tabular}
\end{table}

\clearpage

\clearpage

\subsection{The appendage}
\label{app:design_space_link}

\begin{figure}[!t]
  \centering
  \includegraphics[width=\columnwidth]{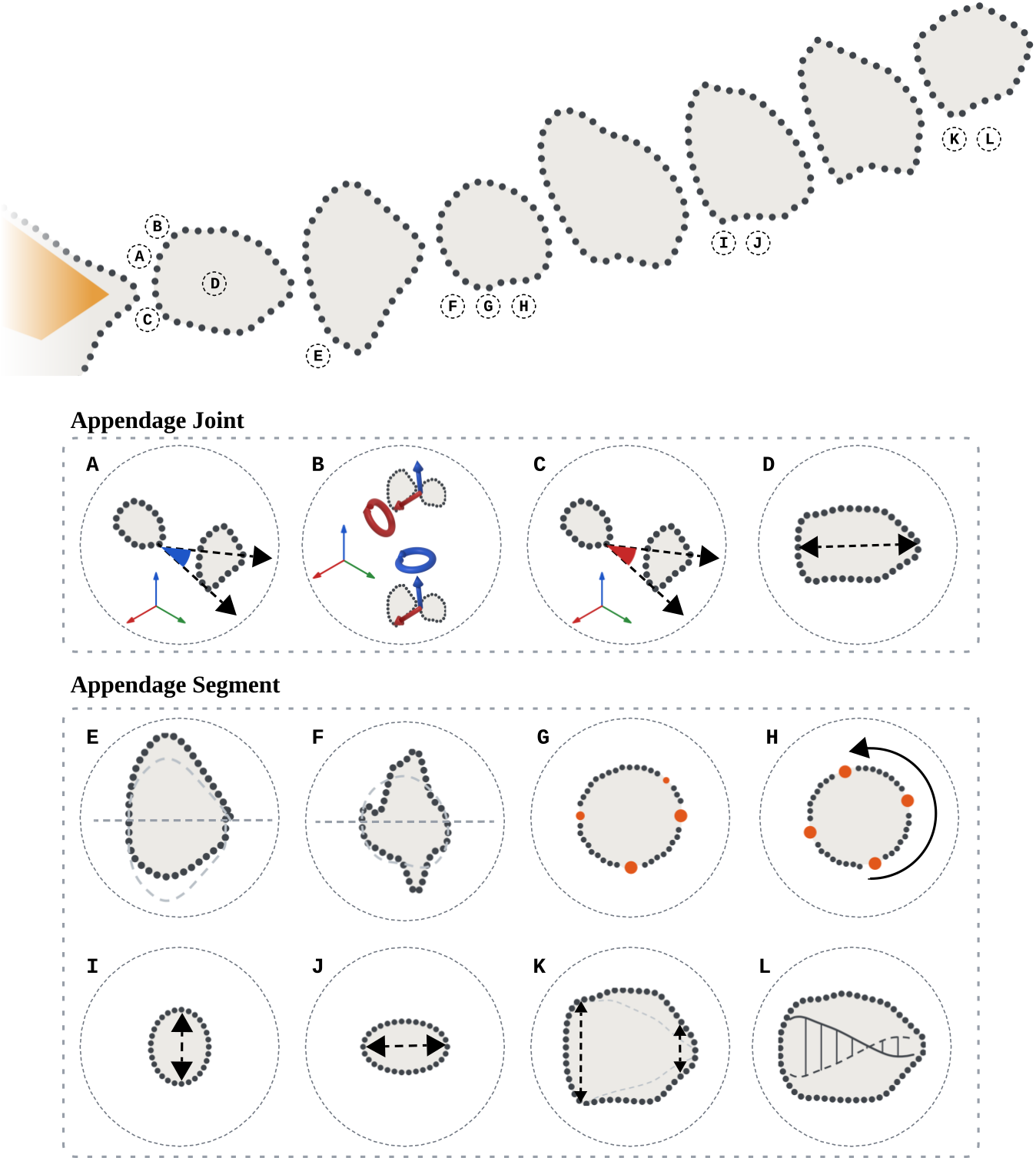}
  \caption{\textbf{The appendage.} Top: one appendage of the same root, growing
from its spoke tip at the left; each segment is drawn as a chain of beads
around its outline. \textbf{A--L}: one parameter each, cut and enlarged from
the appendage above; the small circle marks where each was cut.
Appendage joint: A \textbf{Appendage joint yaw bend},
B \textbf{Appendage joint hinge axis}, C \textbf{Appendage joint pitch bend},
D \textbf{Appendage joint length}.
Appendage segment: E \textbf{Appendage segment asymmetry},
F \textbf{Appendage segment gain}, G \textbf{Appendage segment lobe},
H \textbf{Appendage segment phase}, I \textbf{Appendage segment width},
J \textbf{Appendage segment depth}, K \textbf{Appendage segment taper},
L \textbf{Appendage segment twist}.}

\label{fig:design_space-appendage}

\end{figure}

The appendage has two parts: its segments, the rigid pieces of the chain, and
its joints, one between each segment and the one before it.
The letter after each name is its panel in Fig.~\ref{fig:design_space-appendage}.

Each joint has four parameters.
\begin{list}{}{\setlength{\labelwidth}{5.6cm}\setlength{\labelsep}{0.2cm}\setlength{\leftmargin}{5.8cm}\setlength{\itemsep}{2pt}\setlength{\parsep}{0pt}\setlength{\topsep}{4pt}}
  \item[{\makebox[4.9cm][l]{\textbf{Appendage joint yaw bend}}\makebox[0.7cm][l]{(A)}}]
        turns the segment toward its side, relative to the segment before it, when the joint is at rest.
  \item[{\makebox[4.9cm][l]{\textbf{Appendage joint hinge axis}}\makebox[0.7cm][l]{(B)}}]
        sets which way the axis of the joint runs:
        across the width of the previous segment (lateral), so that the segment folds toward its face,
        or across its depth (normal), so that the segment swings toward its side.
        Each joint chooses separately, so an appendage can fold at one joint and swing at the next.
  \item[{\makebox[4.9cm][l]{\textbf{Appendage joint pitch bend}}\makebox[0.7cm][l]{(C)}}]
        turns the segment toward its face in the same way.
        Together with yaw bend it gives an appendage a built-in curve before any joint moves.
  \item[{\makebox[4.9cm][l]{\textbf{Appendage joint length}}\makebox[0.7cm][l]{(D)}}]
        sets how long the segment is, and so how far this joint lies from the next.
        Long segments reach far with few joints; short segments put more joints into the same reach.
\end{list}

Each segment has eight parameters.
\begin{list}{}{\setlength{\labelwidth}{5.6cm}\setlength{\labelsep}{0.2cm}\setlength{\leftmargin}{5.8cm}\setlength{\itemsep}{2pt}\setlength{\parsep}{0pt}\setlength{\topsep}{4pt}}
  \item[{\makebox[4.9cm][l]{\textbf{Appendage segment asymmetry}}\makebox[0.7cm][l]{(E)}}]
        makes the segment lopsided: the body leans to one side near the joint and to the other side further along.
        At 0 the segment is symmetric.
  \item[{\makebox[4.9cm][l]{\textbf{Appendage segment gain}}\makebox[0.7cm][l]{(F)}}]
        sets how strong the bulges and dents on the surface of the segment are.
  \item[{\makebox[4.9cm][l]{\textbf{Appendage segment lobe}}\makebox[0.7cm][l]{(G)}}]
        adds one bulge about two thirds of the way along the segment and sets its size,
        which gives the segment a pad near its far end.
  \item[{\makebox[4.9cm][l]{\textbf{Appendage segment phase}}\makebox[0.7cm][l]{(H)}}]
        sets where along the segment the bulges and dents fall.
  \item[{\makebox[4.9cm][l]{\textbf{Appendage segment width}}\makebox[0.7cm][l]{(I)}}]
        sets the size of the cross-section from side to side.
  \item[{\makebox[4.9cm][l]{\textbf{Appendage segment depth}}\makebox[0.7cm][l]{(J)}}]
        sets the size of the cross-section from face to face.
        Equal width and depth give a round segment; unequal values give a flattened one.
  \item[{\makebox[4.9cm][l]{\textbf{Appendage segment taper}}\makebox[0.7cm][l]{(K)}}]
        sets how the cross-section changes along the segment:
        below 1 the segment narrows toward its far end, above 1 it widens.
  \item[{\makebox[4.9cm][l]{\textbf{Appendage segment twist}}\makebox[0.7cm][l]{(L)}}]
        rotates the cross-section about the long axis of the segment,
        from none at the joint to the full angle at the far end.
\end{list}

\clearpage

\clearpage
\section{Latent Genome Details}
\label{app:latent-genome}

\subsection{Synthetic data generation}
\label{app:synthetic-dataset}

We generated 2 million designs from the design space described in Appx.~\ref{app:design_space}.
Each design was created in three steps.
First, we chose how many appendages it has, from one to five.
Second, we chose how many joints it has in total, up to 25, and divided them at random among the appendages, 
so that every appendage has at least one joint.
Third, we drew every design parameter of the root and of each segment uniformly at random from its range 
(Table~\ref{tab:design_space_ranges}).

Designs with more appendages have more possible topologies,
because there are more ways to divide the segments among the appendages:
with one appendage there are 25 possible topologies (one to 25 segments), 
and with five appendages there are 1{,}125.
To make sure that the latent space sees every topology about equally often, 
we generated more designs with more appendages, in the ratio 1\,:\,6\,:\,18\,:\,34\,:\,45 
for one to five appendages.
This gave 19{,}240, 115{,}440, 346{,}320, 654{,}000 and 865{,}000 designs, respectively.

\subsection{Encoder}
\label{app:encoder}

The encoder has two branches (Table~\ref{tab:encoder}).
The first branch reads the two-channel voxel grid with a 3D residual network \citep{he2016deep}.
The network shrinks the grid in five steps, from 128 voxels per side to 4,
while the number of channels grows from 32 to 512.
Average pooling then yields a vector of 512 numbers.
The second branch reads the kinematic graph.
Each node is the root or a joint, and stores the location and hinge axis of that joint.
Three rounds of message passing share information between each joint and its neighbors along the appendage, in both directions.
The mean and the maximum over all nodes are then combined into a vector of 128 numbers.
The two vectors are joined, passed through two linear layers, and projected to a vector of 32 numbers.
This vector is rescaled to unit length,
so the latent genome $z$ lies on the unit sphere $S^{31} \subset \mathbb{R}^{32}$.

\subsection{Decoder}
\label{app:decoder}

A design is written as a sequence of tokens (Fig.~\ref{fig:development}H) :
first the 39 variables of the root, 
then the 12 variables of each segment, one appendage after another.
A control token before each segment says whether the segment starts a new appendage or extends the current one,
and a final token ends the sequence.

\begin{figure}[!htb]
    \centering
    \includegraphics[width=\textwidth]{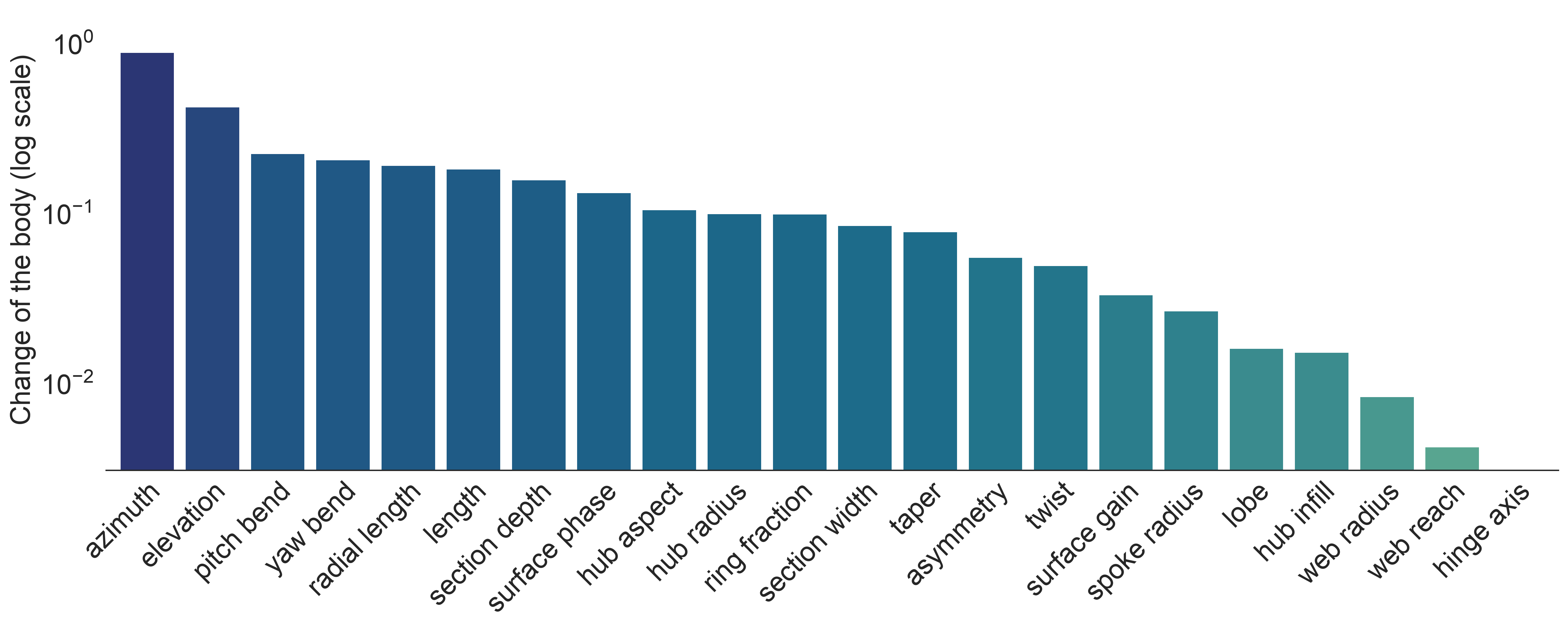}
    \caption{\textbf{Sensitivity analysis of the design variables.}
    We measured how much each variable of the design changes the body.
    Across 1{,}000 designs, each variable in turn was changed by a random amount (10\% of its range on average)
    and the body was voxelized before and after the change.
    The change of the body is one minus the intersection over union (IoU).
    }

    \label{fig:appx_sensitivity_analysis}
\end{figure}

Each continuous variable is cut into equal bins over its range and written as the token of its bin.
We conducted a sensitivity analysis (Fig.~\ref{fig:appx_sensitivity_analysis}) on the parameters of the design in order to decide how many bins each one needs.
Across 1{,}000 designs, 
each variable in turn was changed by a random amount (10\% of its range on average), 
and we measured how much of the voxelized body changed.
Variables that changed the body more were given more bins, from 8 to 128 (Table~\ref{tab:design_space_ranges}).
For example,
the variable that sets where an appendage starts around the edge of the root changed the body the most (86\% of it) and was given 128 bins;
the variables that set the length and the bend of a segment changed 18--22\% of the body and were given 32 bins;
and the variables that set the fine surface detail of a segment changed less than 6\% and were given 8 bins.

The decoder is a transformer \citep{vaswani2017attention} that predicts the next token from the tokens before it (Table~\ref{tab:decoder}).
A linear layer turns the latent genome $z$ into eight vectors, 
and every layer of the transformer attends to them.

\subsection{Training}
\label{app:latent-training}

\begin{figure}[!htb]
\centering
\includegraphics[width=\textwidth]{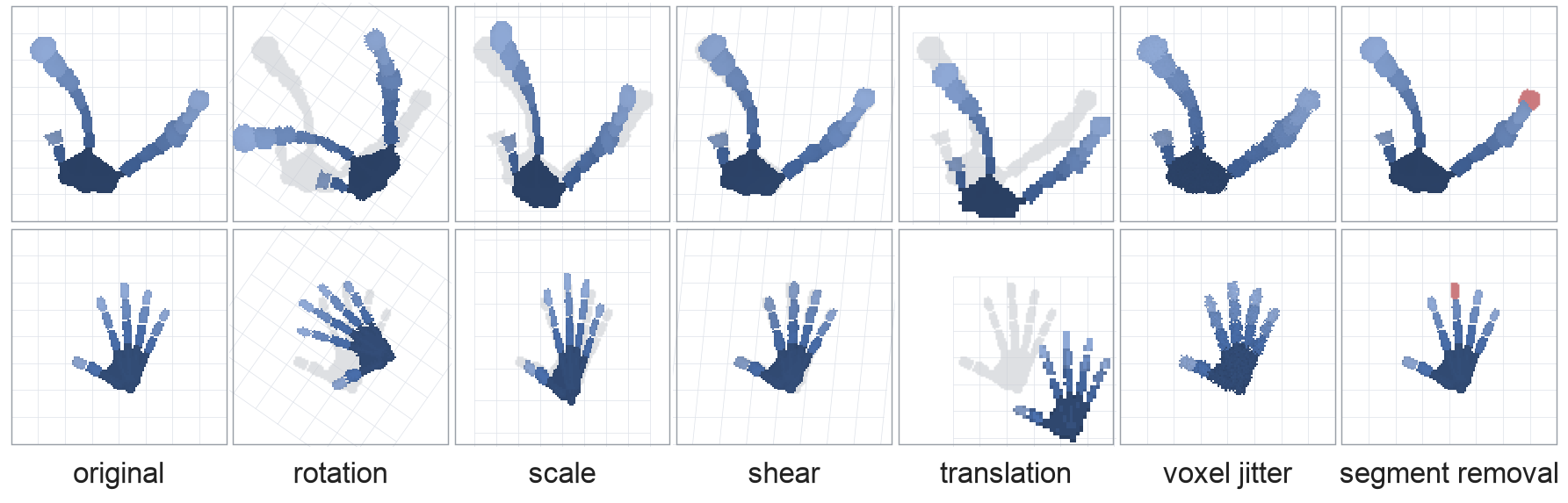}
\caption{\textbf{Data augmentation.}
    Two designs and the operations used to create their positive pairs.
    The original design is shown in gray behind each augmented copy.
    }

\label{fig:appx_augment}
\end{figure}

\paragraph{Positive pairs.}
Each design in a batch is paired with a changed copy of itself, a process known as data augmentation (Fig.~\ref{fig:appx_augment}).
The copy is rotated about the vertical axis of the voxel grid in steps of $30^\circ$,
rescaled by a factor between 0.8 and 1.2 along each axis,
sheared by up to 0.1,
and moved to a random position at which it still fits inside the grid.
With a probability of 0.3, every voxel of the copy is also shifted by up to one voxel.
The same rotation, rescaling, shear and shift are applied to the joints of the kinematic graph.
With a probability of 0.1, the terminal segment of one appendage is removed from the original design.

\paragraph{Loss.}
The encoder and decoder are trained together with a loss that has two terms,
\begin{equation}
    \mathcal{L} = \mathcal{L}_{\mathrm{con}} + \lambda \, \mathcal{L}_{\mathrm{rec}},
\end{equation}
a contrastive term $\mathcal{L}_{\mathrm{con}}$ and a reconstruction term $\mathcal{L}_{\mathrm{rec}}$, 
balanced by the weight $\lambda$.

For the contrastive term, 
a batch of $N$ designs and their $N$ copies is encoded into $2N$ latent genomes, $z_1, \dots, z_{2N}$.
For a design $i$ and its copy $j$, the NT-Xent loss \citep{pmlr-v119-chen20j} is
\begin{equation}
    \ell_{i,j} = -\log \frac{\exp(z_i^{\top} z_j / \tau)}{\sum_{k \neq i} \exp(z_i^{\top} z_k / \tau)},
\end{equation}
where $\tau$ is a temperature.
Because every $z$ has unit length, $z_i^{\top} z_j$ is the cosine similarity of the two latent genomes.
$\mathcal{L}_{\mathrm{con}}$ is the average of $\ell_{i,j}$ and $\ell_{j,i}$ over the $N$ pairs.

For the reconstruction term,
the decoder reads the latent genome of the original design and, at each position $t$ of its token sequence $s_1, \dots, s_T$,
predicts a distribution $\hat{s}_t$ over the next token given the true tokens before it.
The reconstruction term is
\begin{equation}
    \mathcal{L}_{\mathrm{rec}} = -\frac{1}{T} \sum_{t=1}^{T} \log \hat{s}_t[s_t],
\end{equation}
averaged over the batch, where $\hat{s}_t[s_t]$ is the probability given to the true token $s_t$.
The reconstruction term updates the decoder and also the encoder,
so that designs that are close in latent space can still be told apart by the decoder.

\paragraph{Implementation details.}
We set the temperature to $\tau = 0.5$ and the weight of the reconstruction term to $\lambda = 1$.
The model was trained for 50 epochs (31{,}200 steps) with AdamW and a weight decay of $10^{-6}$.
The learning rate rose linearly to $10^{-3}$ over the first five epochs and then followed a cosine decay to zero.
Gradients were clipped to a norm of 10, and training used bfloat16 precision.
Training ran on eight NVIDIA B300 GPUs with a batch of $N = 400$ designs per GPU, and took 16 hours.

\subsection{Evolution}
\label{app:latent-evolution}

Our genes lie on the unit sphere $\mathbb{S}^{31}$, but CMA-ES samples from a normal distribution, which lives in $\mathbb{R}^{32}$.
We therefore let CMA-ES search in $\mathbb{R}^{32}$ and project every sample $\mathbf{x}$ onto the sphere before decoding it,
\begin{equation}
\mathbf{z} = \frac{\mathbf{x}}{\lVert \mathbf{x} \rVert} \in \mathbb{S}^{31}.
\end{equation}
The search starts from a random point on the sphere.
Since only the direction of a sample matters, the length of the distribution's mean carries no information and is free to drift.
The two ablations without the spherical projection have no sphere to search on, so their samples are decoded as they are,
the search starts at the mean of the training genes, and the initial step size is scaled by their standard deviation.

\section{Task Details}
\label{app:tasks-details-entire-section}

All tasks were simulated in Isaac Lab \citep{mittal2025isaac}.
For every design and every task, a control policy was trained from scratch 
with PPO \citep{schulman2017proximal} across thousands of parallel environments,
using a recurrent policy and value function (Table~\ref{tab:ppo}).
To speed up training, we chose to use privileged observations in simulation.
For deployment on the physical robot, the policy of the winning design was first distilled into a proprioceptive policy
and then finetuned under domain randomization (Table.~\ref{tab:domain-randomization}). Below, we describe the setup of each task and the reward it uses and the symbols in the reward tables are defined in Table~\ref{tab:task-symbols}.

\begin{table}[!h]
\centering
\caption{Symbols used in the reward tables.}
\label{tab:task-symbols}
\begin{tabular}{|c|p{0.78\textwidth}|}
\hline
\textbf{Symbol} & \textbf{Meaning} \\
\hline
$\mathbf{p}$, $\mathbf{p}_0$ & position of the object, and its start position \\
$\mathbf{p}^{\ast}$ & point under the root at which the object is held \\
$r$ & distance of the object from its start position along the table, $|\mathbf{p}_{xy} - \mathbf{p}_{0,xy}|$ \\
$h$ & height of the object above its start position \\
$\theta$ & angle between the orientation of the object and the goal orientation \\
$\omega$ & rotation rate of the object about the task axis \\
$\phi$ & angle between the actual and the intended rolling axis of the object \\
$v$ & linear speed of the object \\
$g$, $g^{\star}$ & progress of the object toward the opening of the pipe, from 0 to 1, and its largest value so far \\
$p$ & proximity of the nearest link to the object \\
$s$ & slip: how far the object has moved relative to the root since release \\
$d_{i}$ & distance from the tip of appendage $i$ to the object \\
$f_{i}$ & contact force between the tip of appendage $i$ and the object, clipped to $[0, 1]$\,N \\
$f_{\mathrm{self}}$ & contact force between appendages \\
$\mathbf{a}_t$ & action at step $t$ \\
$\mathbf{q}_t$, $\mathbf{q}_0$ & joint position targets at step $t$, and the initial pose \\
$\boldsymbol{\tau}$ & torque of every joint divided by the torque limit of its motor \\
$\Delta$ & change of a quantity over one step \\
\hline
\end{tabular}
\end{table}

\subsection{Rotate}
\label{app:task-details-rotate}

\begin{wrapfigure}{r}{0.3\textwidth}
  \vspace{-12pt}
  \centering
  \includegraphics[width=\linewidth]{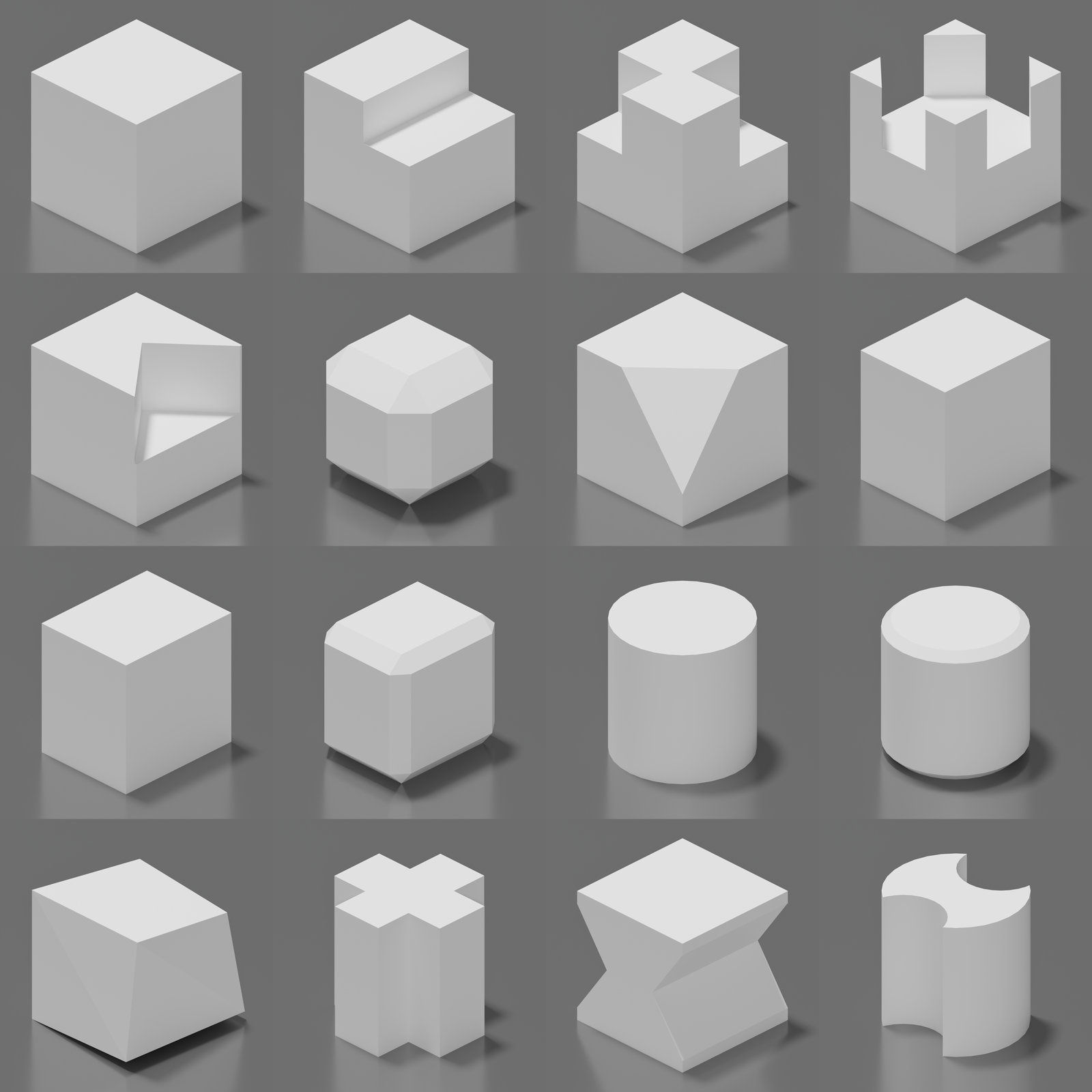}
  \vspace{-16pt}
  \caption{The 16 training objects.}
  \label{fig:training_objects}
  \vspace{-10pt}
\end{wrapfigure}

We trained on 16 objects from \cite{yin2023rotating}, with an equal number of environments per object (Fig.~\ref{fig:training_objects}).
The design is mounted with its root facing upward, and the object is placed on the root.

The policy observes the joint positions, the joint position targets, and the position of the object relative to its start position.
A goal orientation starts at the orientation of the object and advances by $22.5^{\circ}$ about the vertical axis every time it is reached,
so a policy that keeps reaching goals rotates the object continuously.
The reward (Table~\ref{tab:reward-rotate}) follows the goal-based scheme of \cite{handa2023dextreme} and \cite{fay2026house}.
It rewards progress toward the goal and reaching it,
and penalizes the object drifting from where it started.
An episode ends early when the object falls off the root, tips over, or drifts too far.
Fitness is the net rotation rate of the object over a 30 second window, in which nothing ends early.

The policy trained during evolution observes the position of the object, which the physical design cannot sense.
We therefore distilled it into a proprioceptive policy, one that observes only the joint positions and the joint position targets.
The evolution policy acts as a frozen teacher, and the proprioceptive policy is trained to output the action the teacher would take from the same state.
The proprioceptive policy was then fine-tuned further with PPO under the domain randomization listed in Table~\ref{tab:domain-randomization}.
Following \cite{andrychowicz2020learning}, the randomization starts at zero and is raised, step by step, each time the policy has adapted to the current level.

\begin{table*}[!h]
\centering

\def\rotcolwidth{3.8em}

\def\rotvalue#1#2{%
  \makebox[\rotcolwidth][c]{%
    \ensuremath{\text{#1}_{\scriptscriptstyle \pm\text{#2}}}%
  }%
}
\def\rotbold#1#2{%
  \makebox[\rotcolwidth][c]{%
    \ensuremath{\text{\textbf{#1}}_{\scriptscriptstyle
    \boldsymbol{\pm}\text{\textbf{#2}}}}%
  }%
}

\def\rotheader#1#2{%
  \multicolumn{2}{c}{%
    \parbox[c]{\dimexpr\rotcolwidth+\rotcolwidth+2\tabcolsep\relax}{%
      \centering
      \textbf{#1}\par
      (#2 g)%
    }%
  }%
}

\caption{
Real-world in-hand rotation performance across diverse unseen objects.
We report the mean rotation speed ($\omega$, rad/s) and stable rotation time ($T$, s), with sample standard deviations shown as subscripts.
Rotation speeds are evaluated in two trials.
Stable rotation time is evaluated in three 30\,s trials.
For each object, both metrics are bolded for the design with the highest mean rotation speed, determined before rounding.
An em dash indicates failure to sustain in-hand rotation.
}
\label{tab:real_world_rotation}
\vspace{4pt}

\scriptsize
\setlength{\tabcolsep}{1.5pt}
\renewcommand{\arraystretch}{1.15}

\resizebox{\textwidth}{!}{%
\begin{tabular}{@{}l*{10}{c}@{}}

\toprule
&
\rotheader{Wooden Cube}{134.5} &
\rotheader{Rubber Cube}{26.8} &
\rotheader{Rubik's Cube 3x}{92.9} &
\rotheader{Rubik's Cube 4x}{132.3} &
\rotheader{Fabric Cube}{16.1} \\

\cmidrule(lr){2-3}
\cmidrule(lr){4-5}
\cmidrule(lr){6-7}
\cmidrule(lr){8-9}
\cmidrule(lr){10-11}

\textbf{Design}
& $\omega$ & $T$
& $\omega$ & $T$
& $\omega$ & $T$
& $\omega$ & $T$
& $\omega$ & $T$ \\
\midrule

LEAP (Multi-obj)
& \multicolumn{2}{c}{\textemdash}
& \multicolumn{2}{c}{\textemdash}
& \multicolumn{2}{c}{\textemdash}
& \multicolumn{2}{c}{\textemdash}
& \multicolumn{2}{c}{\textemdash} \\

LEAP (Cube-only)
& \rotvalue{1.7}{0.1} & \rotvalue{20.0}{7.0}
& \rotvalue{0.2}{0.0} & \rotvalue{30.0}{0.0}
& \rotvalue{0.3}{0.0} & \rotvalue{30.0}{0.0}
& \rotvalue{0.3}{0.0} & \rotvalue{30.0}{0.0}
& \multicolumn{2}{c}{\textemdash} \\

Ours (Tetradactyl)
& \rotbold{1.8}{0.1} & \rotbold{30.0}{0.0}
& \rotvalue{1.9}{0.0} & \rotvalue{30.0}{0.0}
& \rotvalue{1.7}{0.1} & \rotvalue{30.0}{0.0}
& \rotvalue{1.8}{0.3} & \rotvalue{21.3}{7.0}
& \rotbold{1.7}{0.1} & \rotbold{30.0}{0.0} \\

Ours (Didactyl)
& \rotvalue{1.5}{0.3} & \rotvalue{30.0}{0.0}
& \rotbold{2.4}{0.5} & \rotbold{30.0}{0.0}
& \rotbold{2.0}{0.0} & \rotbold{30.0}{0.0}
& \rotbold{1.9}{0.2} & \rotbold{30.0}{0.0}
& \rotvalue{1.4}{0.2} & \rotvalue{30.0}{0.0} \\

\specialrule{\heavyrulewidth}{5pt}{5pt}

&
\rotheader{Tomato Soup Can}{353.2} &
\rotheader{Moon Ball}{30.9} &
\rotheader{Apple}{364.6} &
\rotheader{Orange}{212.6} &
\rotheader{Avocado}{223.8} \\

\cmidrule(lr){2-3}
\cmidrule(lr){4-5}
\cmidrule(lr){6-7}
\cmidrule(lr){8-9}
\cmidrule(lr){10-11}

\textbf{Design}
& $\omega$ & $T$
& $\omega$ & $T$
& $\omega$ & $T$
& $\omega$ & $T$
& $\omega$ & $T$ \\
\midrule

LEAP (Multi-obj)
& \multicolumn{2}{c}{\textemdash}
& \multicolumn{2}{c}{\textemdash}
& \multicolumn{2}{c}{\textemdash}
& \rotvalue{0.2}{0.0} & \rotvalue{30.0}{0.0}
& \multicolumn{2}{c}{\textemdash} \\

LEAP (Cube-only)
& \multicolumn{2}{c}{\textemdash}
& \multicolumn{2}{c}{\textemdash}
& \rotvalue{0.3}{0.0} & \rotvalue{30.0}{0.0}
& \rotvalue{0.6}{0.0} & \rotvalue{30.0}{0.0}
& \multicolumn{2}{c}{\textemdash} \\

Ours (Tetradactyl)
& \multicolumn{2}{c}{\textemdash}
& \multicolumn{2}{c}{\textemdash}
& \multicolumn{2}{c}{\textemdash}
& \rotvalue{0.9}{0.1} & \rotvalue{30.0}{0.0}
& \multicolumn{2}{c}{\textemdash} \\

Ours (Didactyl)
& \rotbold{1.0}{0.3} & \rotbold{30.0}{0.0}
& \rotbold{1.3}{0.1} & \rotbold{26.8}{2.9}
& \rotbold{1.3}{0.1} & \rotbold{24.7}{1.5}
& \rotbold{1.8}{0.1} & \rotbold{30.0}{0.0}
& \rotbold{1.4}{0.1} & \rotbold{21.7}{4.9} \\

\specialrule{\heavyrulewidth}{5pt}{5pt}

&
\rotheader{Disk Toy}{77.3} &
\rotheader{Hand Soap}{315.6} &
\rotheader{Salt Container}{687.3} &
\rotheader{Jam Jar}{577.7} &
\rotheader{Energy Stick}{68.7} \\

\cmidrule(lr){2-3}
\cmidrule(lr){4-5}
\cmidrule(lr){6-7}
\cmidrule(lr){8-9}
\cmidrule(lr){10-11}

\textbf{Design}
& $\omega$ & $T$
& $\omega$ & $T$
& $\omega$ & $T$
& $\omega$ & $T$
& $\omega$ & $T$ \\
\midrule

LEAP (Multi-obj)
& \multicolumn{2}{c}{\textemdash}
& \multicolumn{2}{c}{\textemdash}
& \multicolumn{2}{c}{\textemdash}
& \multicolumn{2}{c}{\textemdash}
& \multicolumn{2}{c}{\textemdash} \\

LEAP (Cube-only)
& \multicolumn{2}{c}{\textemdash}
& \multicolumn{2}{c}{\textemdash}
& \rotvalue{0.2}{0.0} & \rotvalue{30.0}{0.0}
& \multicolumn{2}{c}{\textemdash}
& \multicolumn{2}{c}{\textemdash} \\

Ours (Tetradactyl)
& \multicolumn{2}{c}{\textemdash}
& \multicolumn{2}{c}{\textemdash}
& \multicolumn{2}{c}{\textemdash}
& \multicolumn{2}{c}{\textemdash}
& \multicolumn{2}{c}{\textemdash} \\

Ours (Didactyl)
& \rotbold{1.0}{0.4} & \rotbold{30.0}{0.0}
& \rotbold{0.8}{0.0} & \rotbold{30.0}{0.0}
& \rotbold{0.9}{0.1} & \rotbold{30.0}{0.0}
& \rotbold{0.6}{0.1} & \rotbold{30.0}{0.0}
& \rotbold{1.6}{0.1} & \rotbold{26.2}{2.7} \\

\bottomrule
\end{tabular}%
}

\end{table*}

\begin{table}[!h]
\centering
\caption{Reward terms of the rotate task.}
\label{tab:reward-rotate}
\begin{tabular}{|l|l|r|}
\hline
\textbf{Term} & \textbf{Equation} & \textbf{Weight} \\
\hline
rotation toward goal & $1 / (|\theta| + 0.1)$ & $1$ \\
goal reached & $|\theta| \le 0.2$ & $250$ \\
steady rotation & $0.25 < \omega_z < 1.5$ & $1$ \\
object displacement & $|\mathbf{p} - \mathbf{p}_0|$ & $-10$ \\
object drift & $\max(|\mathbf{p} - \mathbf{p}_0| - 0.03,\, 0)$ & $-0.1$ \\
action magnitude & $|\mathbf{a}_t|^2$ & $-2\mathrm{e}{-4}$ \\
pose deviation & $|\mathbf{q}_t - \mathbf{q}_0|^2$ & $-0.1$ \\
drop & episode ends & $-10$ \\
\hline
\end{tabular}
\end{table}

We then deployed the policy zero-shot on the physical design and tested 15 objects that were not used in training (Table~\ref{tab:real_world_rotation}).
A design that could not keep an object rotating is marked with a dash.
The didactyl design rotated every object, at 1.39 rad/s on average,
and sustained rotation for the full 30 seconds on 11 of them.
The tetradactyl design rotated the five cubes and the orange but no other object.
On the cubes it was as fast as the didactyl design, at 1.78 rad/s on average.
The LEAP Hand trained on all 16 objects failed to rotate any of the test objects, so we trained a second policy for it on a single cube.
That policy rotated the cubes and three other objects, at 0.51 rad/s on average, a third of the speed of the didactyl design.

\subsection{Pick}
\label{app:task-details-pick}

We trained on the same 16 objects as in the rotate task.
The design is mounted above a table with its root facing downward, and the object is placed on the table beneath it.
The object counts as held when it is lifted off the table and kept under the root.
The reward (Table~\ref{tab:reward-pick}) follows the usual shape for learning to grasp \citep{wan2023unidexgrasp++,yuan2025cross}.
It pays for approaching and touching the object, for lifting it up to the root, and for holding it still there,
and penalizes the object drifting sideways or moving fast.
An episode ends early when a held object falls back to the table or escapes from under the root.
Fitness is how long the object stays held over a 60 second window in which a force pulls on it,
a harder form of the test used to validate grasps in \citet{wang2022dexgraspnet}.
The direction of the force changes at random every 3 seconds, and its strength increases every 3 seconds.

\begin{table}[!h]
\centering
\caption{Reward terms of the pick task.}
\label{tab:reward-pick}
\begin{tabular}{|l|l|r|}
\hline
\textbf{Term} & \textbf{Equation} & \textbf{Weight} \\
\hline
approach & $\mathrm{mean}_{i}\, \exp(-d_{i} / 0.02)$, before first contact & $0.4$ \\
contact & $\mathrm{mean}_{i}\, f_{i}$ & $0.8$ \\
multiple contacts & at least two appendages touch the object & $0.8$ \\
lift & $\mathrm{clip}\big((h - 0.022) / 0.008,\, 0,\, 1\big)$ & $8$ \\
near the root & $\exp(-|\mathbf{p} - \mathbf{p}^{\ast}| / 0.05)$ & $2$ \\
stable grasp & held, at least two contacts, and nearly at rest & $2$ \\
sideways drift & $\max(r - 0.04,\, 0)$ & $-5$ \\
linear speed & $v$ & $-0.25$ \\
angular speed & $\omega$ & $-0.02$ \\
action magnitude & $|\mathbf{a}_t|^2$ & $-2\mathrm{e}{-4}$ \\
self-collision & $f_{\mathrm{self}}$ & $-0.02$ \\
drop & object dropped & $-10$ \\
\hline
\end{tabular}
\end{table}

\begin{table*}[!h]
\centering
\caption{
Real-world performance on the pick task across diverse objects.
Results indicate successful picks out of 10 trials per object.
}
\label{tab:real_world_grasping}
\vspace{2pt}

\begingroup
\footnotesize
\setlength{\tabcolsep}{2pt}
\renewcommand{\arraystretch}{1.05}

\newcommand{\graspobject}[1]{%
  \parbox[c]{\dimexpr(\textwidth-22\tabcolsep)/12\relax}{%
    \centering
    \scriptsize
    \textbf{#1}\strut\par
  }%
}

\begin{tabular}{@{}*{12}{c}@{}}
\toprule

\graspobject{Wooden\\Cube} &
\graspobject{Rubber\\Cube} &
\graspobject{Rubik's\\Cube 3x} &
\graspobject{Rubik's\\Cube 4x} &
\graspobject{Fabric\\Cube} &
\graspobject{Cube\\Toy} &
\graspobject{Lemonade\\Mix Box} &
\graspobject{Tennis\\Ball} &
\graspobject{Moon\\Ball} &
\graspobject{Orange} &
\graspobject{Avocado} &
\graspobject{Donut\\Toy} \\

\midrule

10/10 & 10/10 & 8/10 & 8/10 & 10/10 & 7/10 &
10/10 & 6/10 & 5/10 & 9/10 & 5/10 & 4/10 \\

\midrule

\graspobject{Water\\Can} &
\graspobject{Candle\\Jar} &
\graspobject{Pudding} &
\graspobject{Polygon\\Toy} &
\graspobject{Jam\\Jar} &
\graspobject{Energy\\Stick} &
\graspobject{Meat\\Can} &
\graspobject{Toy\\Bear} &
\graspobject{Disney\\Toy} &
\graspobject{Baby\\Bottle} &
\graspobject{Motor} &
\graspobject{Salt\\Container} \\

\midrule

10/10 & 6/10 & 10/10 & 8/10 & 6/10 & 10/10 &
10/10 & 10/10 & 10/10 & 10/10 & 10/10 & 0/10 \\

\bottomrule
\end{tabular}
\endgroup

\end{table*}

The policy was distilled into a proprioceptive policy and fine-tuned as in the rotate task.
We then deployed the policy zero-shot on the physical design and tested 24 objects, with 10 attempts each (Table~\ref{tab:real_world_grasping}).
In each attempt the design was lowered onto the object, closed its appendages, and was lifted.
An attempt counts as a success if the object was lifted off the table and stayed in the hand.
The design picked up 23 of the 24 objects at least once and succeeded in 80\% of all attempts.
It failed only on the salt container, the heaviest object, and was least reliable on round objects such as the moon ball and the avocado.

\subsection{Hang}
\label{app:task-details-hang}

The design starts next to a horizontal bar, with a 500\,g weight hanging below its root.
Each episode has two phases.
In the first, gravity is off and the root is held in place while the design closes its appendages around the bar.
Then the root is released and gravity is switched on, so the design must carry its own weight and the load.
Throughout, the bar twists back and forth about its axis.
The design counts as holding on when at least two appendages touch the bar and the root has not slipped away from it.
The reward (Table~\ref{tab:reward-hang}) pays for approaching and touching the bar and, after release, for holding on without slipping,
and penalizes the torque the joints use.
An episode ends early when the design loses contact with the bar, slips too far, or falls.
Fitness is the time the design held on over a 20 second window in which the bar twists twice as far as in training.
Each second held counts for less the more torque the joints used, down to half.

We deployed the policy zero-shot on the physical design, which gripped a stationary bar and hung from it under the 500\,g load (Fig.~\ref{fig:robot_gallery_pick_hang_retrieve}E-H).

\begin{table}[!h]
\centering
\caption{Reward terms of the hang task.
Terms marked with $\dagger$ are active only after release. $\tau_{\mathrm{ref}}$ is a hyperparameter.}
\label{tab:reward-hang}
\begin{tabular}{|l|l|r|}
\hline
\textbf{Term} & \textbf{Equation} & \textbf{Weight} \\
\hline
approach & $\mathrm{mean}_{i}\, \exp(-d_{i} / 0.02)$, before first contact & $0.4$ \\
contact & $\mathrm{mean}_{i}\, f_{i}$ & $0.8$ \\
multiple contacts & at least two appendages touch the bar & $0.8$ \\
stability$^{\dagger}$ & $\mathrm{mean}_{i}\, f_{i} \cdot \exp(-s / 0.0075)$, if at least two contacts & $8$ \\
stable grasp$^{\dagger}$ & at least two contacts and $s \le 0.0075$ & $2$ \\
joint torques & $|\boldsymbol{\tau}|^2 / \tau_{\mathrm{ref}}$ & $-0.5$ \\
action smoothing & $|\mathbf{a}_t - \mathbf{a}_{t-1}|^2$ & $-0.001$ \\
self-collision & $f_{\mathrm{self}}$ & $-0.02$ \\
failure & episode ends & $-10$ \\
\hline
\end{tabular}
\end{table}

\subsection{Retrieve}
\label{app:task-details-retrieve}

A cube rests inside a narrow horizontal pipe.
The design is mounted at the opening of the pipe with its root fixed, so only its appendages can enter.
It must reach in, hook the cube from behind, and drag it out.
Every episode starts from a pose, searched for before training, in which one link reaches behind the cube.
The policy observes the joint positions, the joint position targets, and the position of the cube inside the pipe.
The reward (Table~\ref{tab:reward-retrieve}) pays for moving the cube toward the opening, with bonuses for passing milestones along the way and for getting it out.
An episode ends when the cube is out of the pipe.

\begin{table}[!h]
\centering
\caption{Reward terms of the retrieve task.}
\label{tab:reward-retrieve}
\begin{tabular}{|l|l|r|}
\hline
\textbf{Term} & \textbf{Equation} & \textbf{Weight} \\
\hline
new progress & $\min(\Delta g^{\star},\, 0.1)$ & $40$ \\
new progress in contact & $\min(\Delta g^{\star},\, 0.1)$ & $40$ \\
new progress, unclipped & $\Delta g^{\star}$ & $80$ \\
new progress, squared & $\Delta (g^{\star})^{2}$ & $80$ \\
net progress & $\mathrm{clip}(\Delta g,\, \pm 0.1)$ & $10$ \\
proximity & $\mathrm{clip}(\Delta p,\, \pm 0.25)$ & $5$ \\
contact & in contact & $0.005$ \\
milestones & $g^{\star}$ passes $0.25,\, 0.5,\, 0.65,\, 0.8,\, 0.95$ & $2,\, 4,\, 8,\, 12,\, 20$ \\
retrieved & retrieved & $400$ \\
escape & escaped & $-120$ \\
action smoothing & $|\mathbf{a}_t - \mathbf{a}_{t-1}|^2$ & $-0.001$ \\
\hline
\end{tabular}
\end{table}

Most designs cannot move the cube at all, so a fitness that only counted retrievals would be zero for almost every design.
Fitness therefore ranks designs in stages.
A design that cannot touch the cube is scored by how close it gets,
one that touches it by how far behind it it reaches,
and one that hooks it by how far it can drag it.
Only designs that can drag the cube are trained, and they are scored above all others by the fraction of episodes in which they retrieve it.
The policy was distilled into a proprioceptive policy and fine-tuned, then deployed zero-shot on the physical design, it retrieved the cube from a real pipe (Fig.~\ref{fig:robot_gallery_pick_hang_retrieve}I-K).

\subsection{Flip}
\label{app:task-details-flip}

A cube rests on a table, and the design is mounted above it with its root facing downward.
The design must flip the cube over its edge, again and again, in one fixed direction.
It cannot hold the cube in the air, and must use the table, gravity, and its own body together to turn it,
which is known as extrinsic dexterity \citep{dafle2014extrinsic,zhou2023learning}.
Each flip moves the cube one edge length along the table, so the design must also pull it back to keep flipping it.
The policy observes the joint positions, the joint position targets, the position and rotation rate of the cube, and the current stage.
Each episode alternates between two stages.
In the flip stage, a goal orientation advances by $22.5^{\circ}$ in the rolling direction every time it is reached,
so a policy that keeps reaching goals rolls the cube continuously.
After four goals, that is, one face, the return stage begins, in which the design must bring the cube back to where it started.
The reward (Table~\ref{tab:reward-flip}) pays for progress toward the goal in the flip stage and toward the start position in the return stage,
and penalizes rolling the cube backward, lifting it, turning it off its rolling axis, or losing contact with it.
An episode ends early when the cube strays too far from its start position, is lifted, or leaves the table.
Fitness is the number of full turns of the cube over a 60 second window.

\begin{table}[!h]
\centering
\caption{Reward terms of the flip task.}
\label{tab:reward-flip}
\begin{tabular}{|l|l|r|}
\hline
\textbf{Term} & \textbf{Equation} & \textbf{Weight} \\
\hline
\multicolumn{3}{|l|}{\emph{Flip stage}} \\
\hline
progress toward goal & $\max(-\Delta\theta,\, 0)$ & $10$ \\
goal reached & $\theta \le 0.1$ & $250$ \\
steady rolling & $0.25 < \omega < 1.5$ & $1$ \\
rolling backward & $\max(-\omega,\, 0)$ & $-1$ \\
\hline
\multicolumn{3}{|l|}{\emph{Return stage}} \\
\hline
progress toward start & $-\Delta \max(r - 0.03,\, 0)$ & $5000$ \\
returned & $r \le 0.03$ & $250$ \\
rolling backward & $\max(-\omega,\, 0)$ & $-5$ \\
\hline
\multicolumn{3}{|l|}{\emph{Both stages}} \\
\hline
displacement & $\max(r - 0.03,\, 0)$ & $-10$ \\
lift & $\max(h - 0.015,\, 0)$ & $-10$ \\
off axis & $1 - \cos\phi$ & $-2$ \\
no contact & no appendage touches the cube & $-1$ \\
action magnitude & $|\mathbf{a}_t|^2$ & $-2\mathrm{e}{-4}$ \\
pose deviation & $|\mathbf{q}_t - \mathbf{q}_0|^2$ & $-0.3$ \\
failure & episode ends & $-10$ \\
\hline
\end{tabular}
\end{table}

\subsection{Hold}
\label{app:task-details-hold}

The design is mounted on the end of a robot arm, and a hammer is held in place next to its root.
Each episode has two phases.
In the first, the design closes its appendages around the handle.
Then the hammer is released, and the arm swings it up and down while a force pulls on it in a direction that changes at random.
The design must therefore hold the hammer so that it stays put while it is used, called functional grasp \citep{agarwal2023dexterous}.
The design counts as holding the hammer when at least two appendages touch the handle.
The policy observes the joint positions, the joint position targets, and the pose and velocity of the hammer relative to the root.
The reward (Table~\ref{tab:reward-hold}) rewards approaching and touching the handle and, after release, holding it without letting it slip along, across, or about the handle.
An episode ends early when the hammer drops or moves too far from the root.
Fitness is how still the hammer stays relative to the root over a 30 second window of swinging.
It is highest when the hammer does not move at all, falls with the speed of the hammer relative to the root and with how far it has shifted from its pose at release, and is zero if the hammer drops.

\begin{table}[!h]
\centering
\caption{Reward terms of the hold task.
The grasp score is $G = 0.2\, \mathrm{mean}_{i}\, e^{-d_{i}/0.02} + 0.4\, \mathrm{mean}_{i}\, f_{i} + 0.4$,
and the slip cost is $S = 0.25\,(s_{\mathrm{a}}/0.02)^2 + 0.25\,(s_{\mathrm{l}}/0.015)^2 + 0.5\,(s_{\mathrm{r}}/0.35)^2$,
where $s_{\mathrm{a}}$, $s_{\mathrm{l}}$ and $s_{\mathrm{r}}$ are the slip along, across, and about the handle.
Terms marked with $\dagger$ are active only after release;
those marked with $\ddagger$ are, in addition, multiplied by $\mathrm{mean}_{i}\, f_{i}$ when at least two appendages touch the handle, and by 0 otherwise.}
\label{tab:reward-hold}
\begin{tabular}{|l|l|r|}
\hline
\textbf{Term} & \textbf{Equation} & \textbf{Weight} \\
\hline
grasp progress & $G_t - G_{t-1}$, before release & $0.5$ \\
contact$^{\dagger}$ & $\mathrm{mean}_{i}\, f_{i}$ & $0.25$ \\
multiple contacts$^{\dagger}$ & at least two appendages touch the handle & $0.25$ \\
hold$^{\ddagger}$ & $1$ & $1$ \\
no slip$^{\ddagger}$ & $1 / (1 + S)$ & $8$ \\
action smoothing & $|\mathbf{a}_t - \mathbf{a}_{t-1}|^2$ & $-0.005$ \\
self-collision & $f_{\mathrm{self}}$ & $-0.02$ \\
drop & hammer dropped & $-10$ \\
\hline
\end{tabular}
\end{table}

\subsection{Multitask}
\label{app:task-details-multitask}

\begin{figure}[!h]
    \centering
    \includegraphics[width=\textwidth]{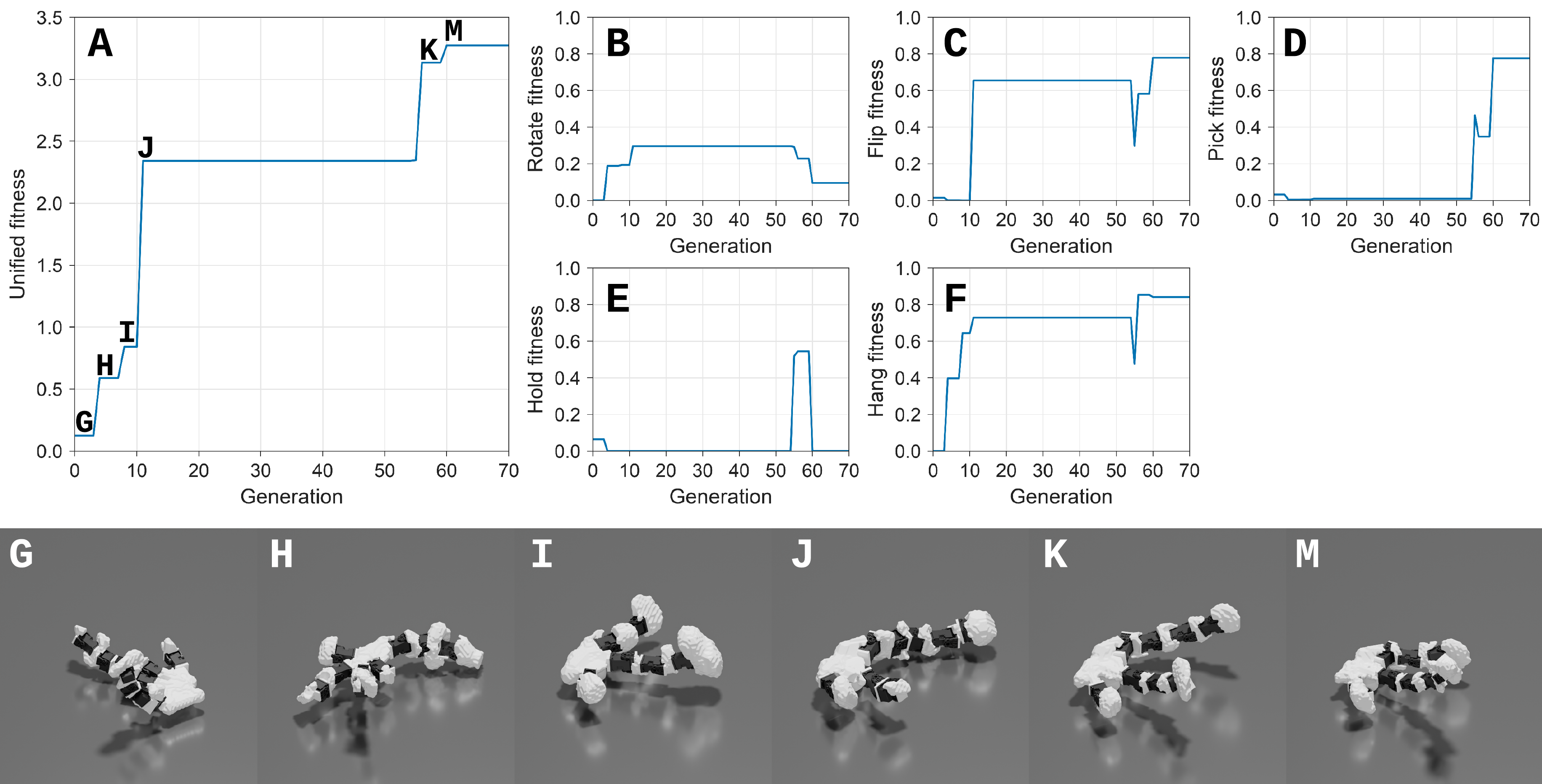}
    \caption{\textbf{The evolution of multitaskers.}
    Robots were evolved to rotate, flip, pick, hold, and hang. Unified fitness (A) was the sum of task normalized scores. The task scores of the best design found so far (B-F) reveal trade-offs: gains in unified fitness did not always improve performance on every task. Selected designs (G-M) correspond to the labeled points in A.
}
    \label{fig:appx_unified_fitness_curve}
\end{figure}

Each design was evaluated in the rotate, flip, pick, hold, and hang environments, with a separate policy trained for each task as in the single-task trials.
The five task scores were divided by the best fitness of the corresponding specialist available when the trial began, then summed to obtain unified fitness.
We ran a single trial of 70 generations (Fig.~\ref{fig:appx_unified_fitness_curve}).

Early improvements came mainly from rotate and hang, followed by a sharp increase in flip fitness around generation 11.
The resulting design remained the leader until generation 55.
It performed well at flipping and hanging but rotated slowly and could neither pick up objects nor hold the hammer.
During this plateau, both unified fitness and the task scores of the leading design remained unchanged.

Around generation 55, designs emerged with nonzero scores on all five tasks, producing a sharp increase in unified fitness.
The design from generation 56 differed from the earlier leader by only two topology edits: one link was removed from its long appendage and another added to its shorter appendage, making the two working appendages more alike.
This change was accompanied by the ability to lift almost every object and hold the hammer through most of the swing, while retaining strong flip and hang performance.
Further gains in unified fitness, however, did not preserve all five abilities.
The highest-scoring design, found in generation 60, improved flip and pick performance but dropped the hammer in almost every episode and rotated more slowly.

To select the final generalist, we divided each task score by the best score reached on that task within this trial and summed again.
This changed the relative weighting of the tasks and ranked the design from generation 56 first.
We selected this design, which achieved nonzero fitness on all five tasks (Fig.~\ref{fig:robot_gallery_generalist}).

Compared with the specialists, the selected generalist flipped the cube more often and hung from the bar nearly as long.
It lifted almost every object but maintained its grasp under the pulling force for only a quarter of the evaluation window, whereas the pick specialist maintained its grasp for nearly the entire window.
It dropped the hammer in one episode in five, compared with one in thirty for the hold specialist, and rotated objects at a quarter of the rate of the rotate specialist.
Rotate and hold showed the largest performance gaps; their specialists also had the most appendages.

\subsection{Real-World Deployment}
\label{app:real-world}

Every joint of a physical design is driven by a Dynamixel XL330-M288-T servo.
The servos are chained on one bus and driven through a U2D2 controller from a computer running ROS2.
Each servo runs its built-in position controller, with the same stiffness and damping as in simulation.
At every step of the 30\,Hz loop.
Figs.~\ref{fig:appx_real_rotation_2finger}-\ref{fig:appx_real_pick} show the didactyl and tetradactyl designs rotating, and the pick design lifting, each of the test objects.


\begingroup
\makeatletter
\setlength{\@fptop}{0pt}
\makeatother

\begin{figure}[!b]
    \centering
    \includegraphics[width=\textwidth]{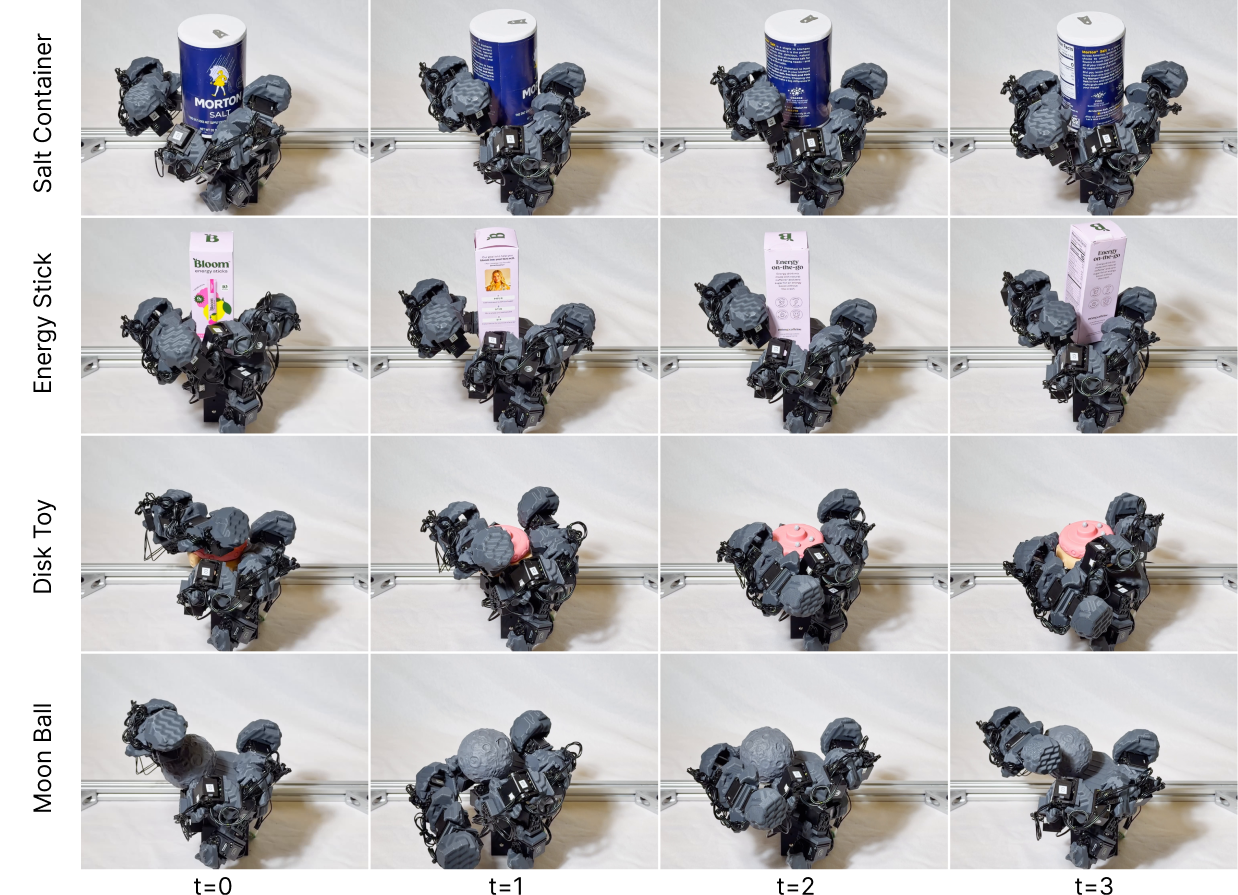}
    \caption{
        \textbf{Real-world in-hand rotation across diverse objects for didactyl design.}
    }
    \label{fig:appx_real_rotation_2finger}
\end{figure}

\begin{figure}[!t]
    \centering
    \includegraphics[width=\textwidth]{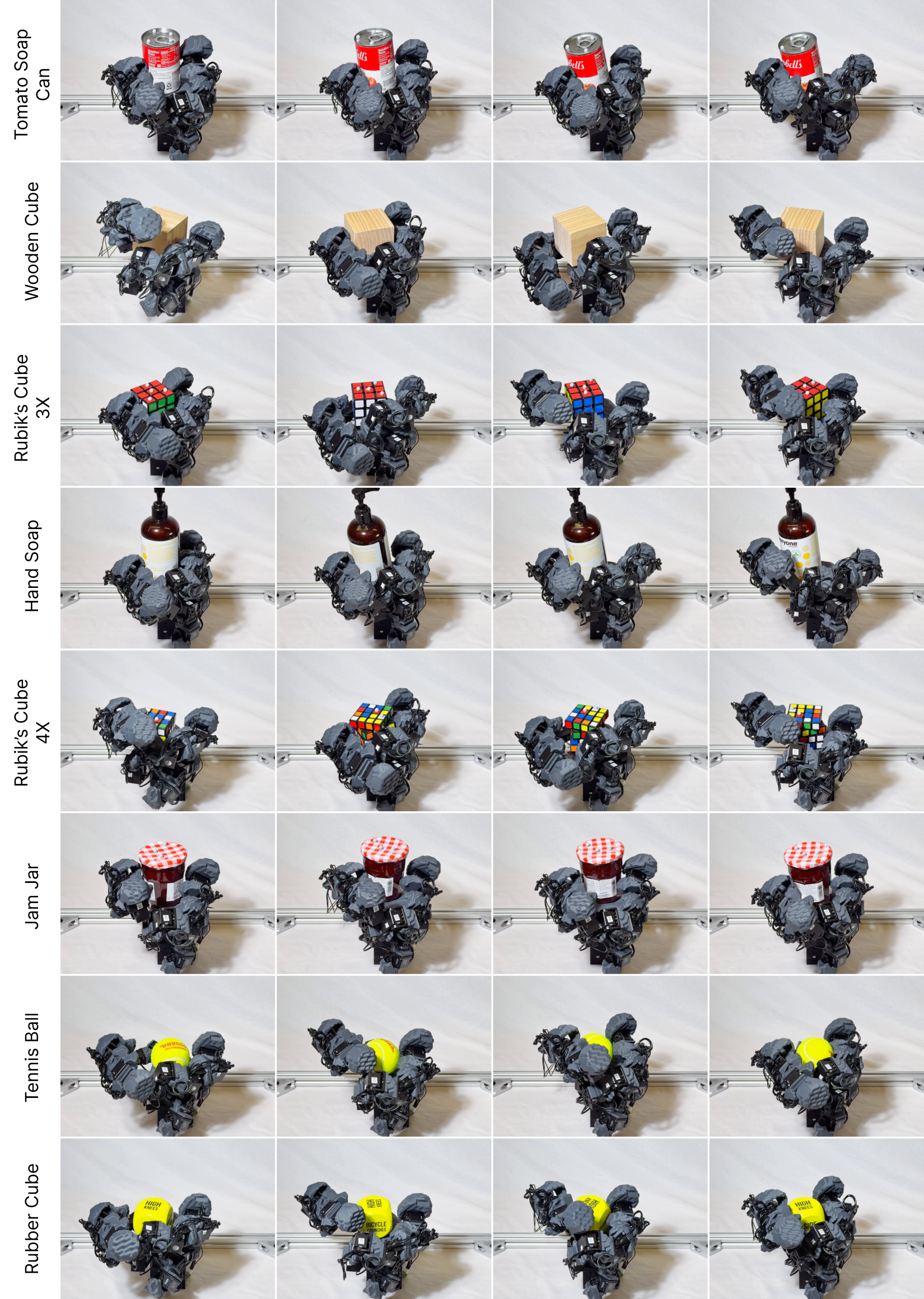}
    \caption{
        \textbf{Real-world in-hand rotation across diverse objects for didactyl design.}
    }
    \label{fig:appx_real_rotation_2finger_2}
\end{figure}

\begin{figure}[!t]
    \centering
    \label{fig:real-rotate-4-appendages}
    \includegraphics[width=\textwidth]{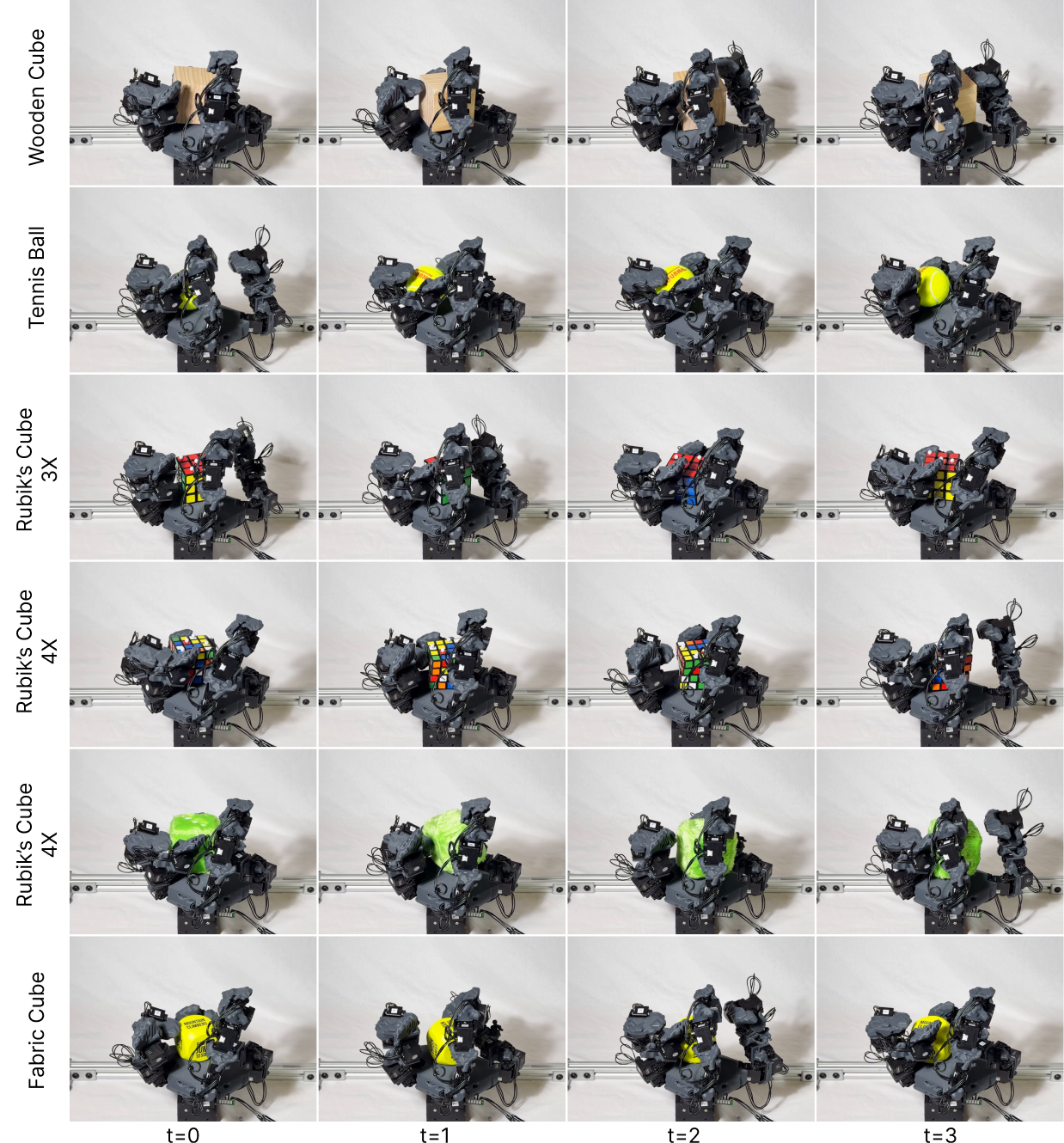}
    \caption{
        \textbf{Real-world in-hand rotation across diverse objects for tetradactyl design.}
    }
    \label{fig:appx_real_rotation_4finger}
\end{figure}

\clearpage

\begin{figure}[!t]
    \centering
    \includegraphics[width=\textwidth]{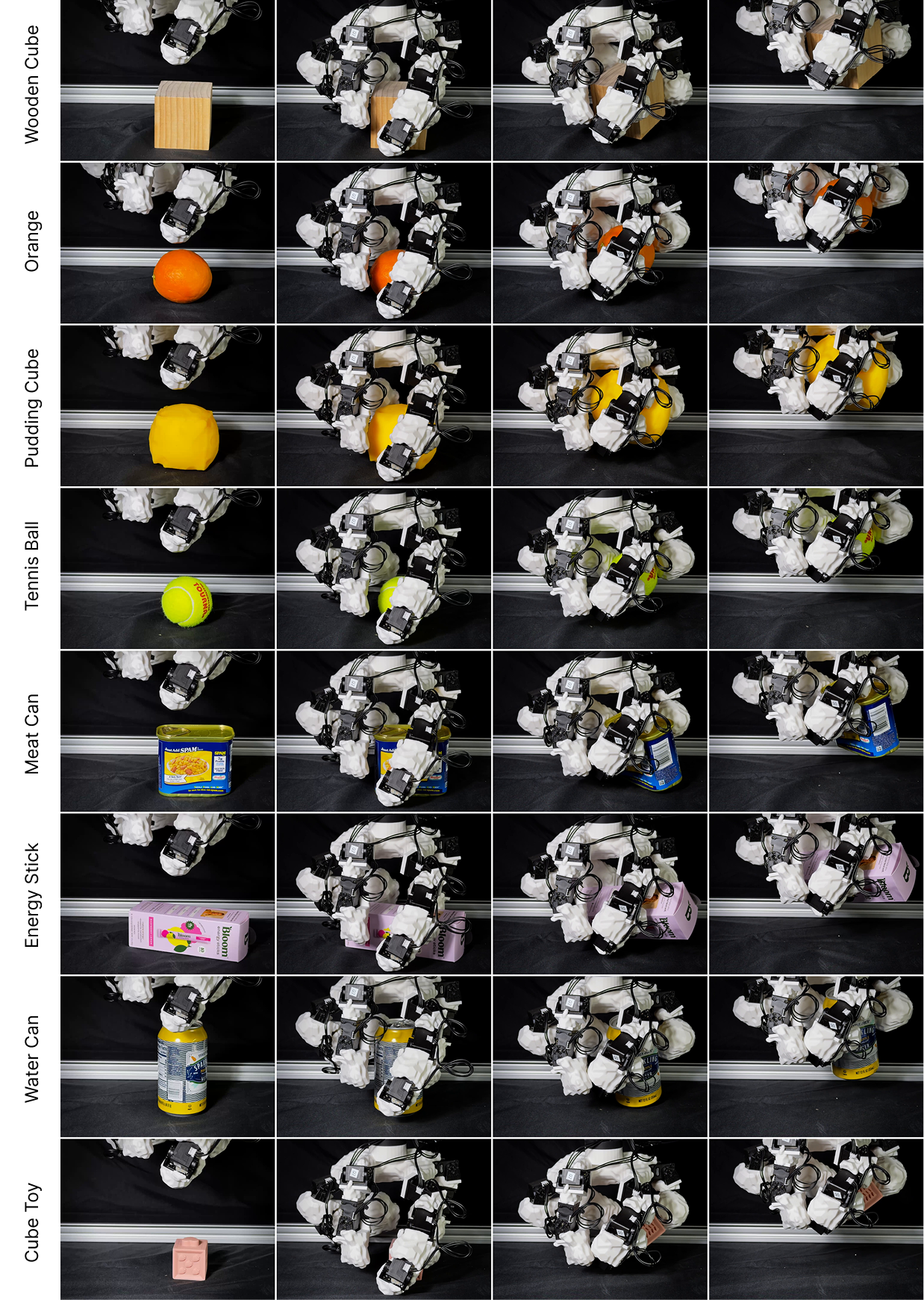}
    \caption{
        \textbf{Real-world picking across diverse objects using the four-appendage design.}
    }
    \label{fig:appx_real_pick2}
\end{figure}

\clearpage

\begingroup
\makeatletter
\setlength{\@fptop}{0pt}
\makeatother

\begin{figure}[!t]
    \centering
    \includegraphics[width=\textwidth]{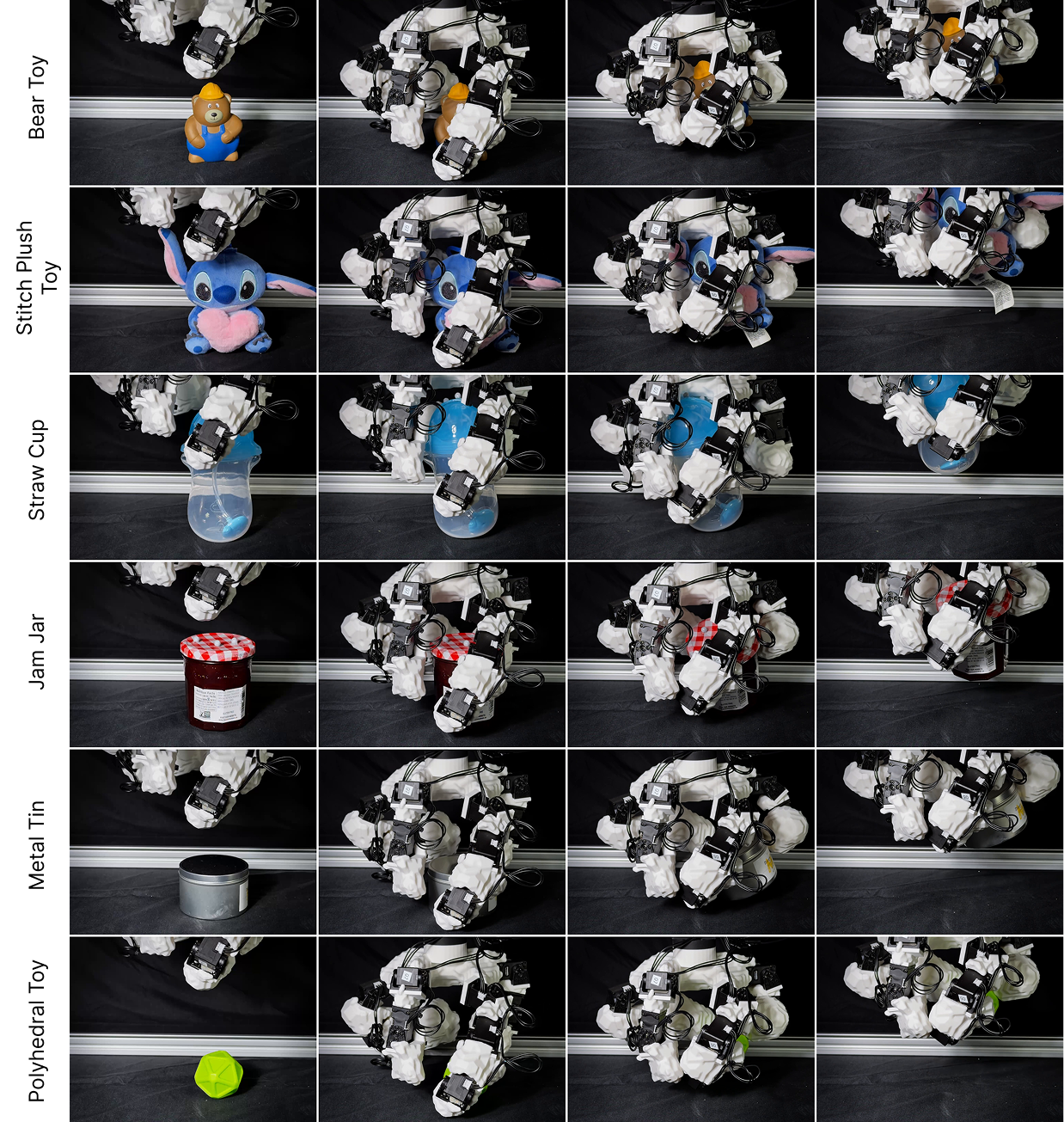}
    \caption{
        \textbf{Real-world picking across diverse objects using the four-appendage design.}
    }
    \label{fig:appx_real_pick}
\end{figure}

\clearpage

\begin{figure}[!t]
  \centering
  \includegraphics[width=\columnwidth]{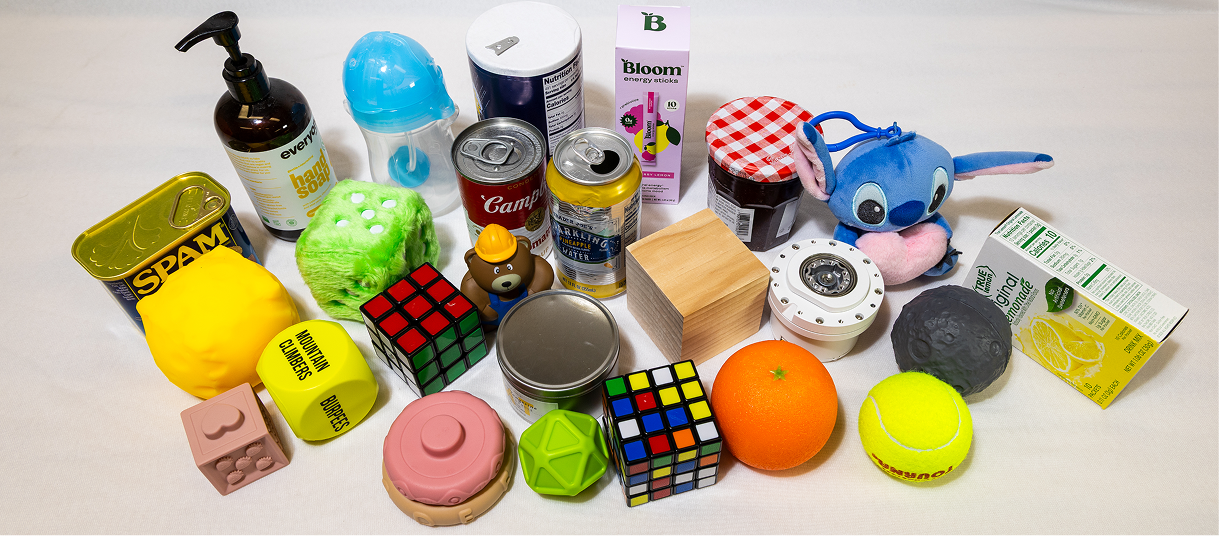}
  \caption{\textbf{Real-world test objects.} The everyday objects used in the
real-world rotate and pick experiments: rigid and soft, smooth and textured,
from a 16~g fabric cube to a 687~g salt container.}

  \label{fig:all_objects}
\end{figure}

\begin{figure}[!t]
  \centering
  \includegraphics[width=\columnwidth]{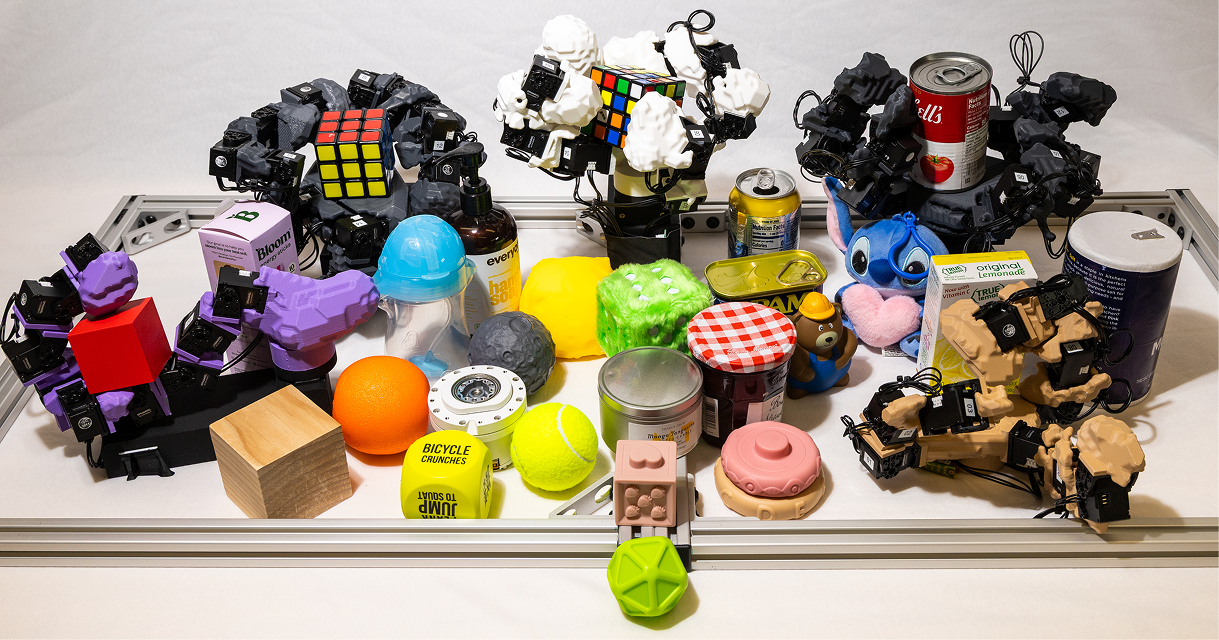}
  \caption{\textbf{The five physical designs.} 
  From left to right: the retriever
(purple), the didactyl rotator (grey), the picker (white), the
4-appendage rotator (grey) and the hanger (tan). Each is 3D-printed in PLA
with one actuator per joint; parts are listed in Table~\ref{tab:bom}. Their real-world results are in
Tables~\ref{tab:real_world_rotation} and~\ref{tab:real_world_grasping} and
Figs.~\ref{fig:robot_gallery_rotation}--\ref{fig:robot_gallery_pick_hang_retrieve}.}

  \label{fig:all_objects_hands}
\end{figure}

\clearpage
\section{Ablation analysis}
\label{app:ablation_analysis}

Our embedding performed well in evolution,
the embedding built without contrastive loss performed worse,
and the embedding generated by the VAE performed far worse (Fig.~\ref{fig:fitness-curves}B).

In this section, we ask why:
what properties of the latent genome produced by contrastive learning allow a black-box optimizer such as CMA-ES to search it efficiently?

CMA-ES samples genes around its current estimate and updates that estimate
using the fitness of the resulting designs.
For this search to be effective, the latent genome must satisfy two requirements:
its genes must decode to useful designs, and distances between genes must
reflect meaningful changes in design.

We define the first requirement as validity.
The gene of an existing design should reconstruct its body plan, while
genes sampled during evolution should decode to designs that pass the
checks required for evaluation.
In evolution, these checks extend from basic geometry to motor fit, mesh
generation, and finding a feasible initial pose.
Failures at different stages receive different fitness values, but samples that
fail cannot be evaluated on the manipulation task.
Here we test reconstruction fidelity and basic geometric validity; These tests probe the requirement, but do not establish that a design will pass the full evaluation
pipeline (see Table~\ref{tab:cmaes_hyperparameters} for the rejection stages
and their fitness values).

We call the second requirement smoothness.
Different poses of the same body plan should map to the same part of the genome,
so that pose does not become an extraneous direction of search.
Increasingly large edits to a body plan should move its gene increasingly far,
and nearby genes should decode to body plans with similar structure and size.
Together, these properties make local sampling more likely to explore related
designs and allow small changes to accumulate during search.


\begin{table}[!h]
\centering
\caption{Ablation of the latent genome.
All validity metrics, Recall@1, and Edit $\rho$ use the
2{,}000-design test set; all three 10-NN metrics use a separate set of
4{,}500 designs.
The \colorbox{bestcol}{best} and \colorbox{secondcol}{second best}
values in each column are shaded.}
\label{tab:latent-ablation}
\footnotesize
\setlength{\tabcolsep}{3pt}
\begin{tabular}{lcccccccc}
\toprule
\multirow{2}{*}{\raisebox{-1ex}{Model}}
& \multicolumn{3}{c}{Validity}
& \multicolumn{5}{c}{Smoothness} \\
\cmidrule(lr){2-4}\cmidrule(lr){5-9}
& App. acc. & Joint acc. & Valid & Recall@1 & Edit $\rho$ &
\makecell{Topology\\10-NN acc.} & \makecell{Length\\10-NN $r$} & \makecell{Volume\\10-NN $r$} \\
\midrule
Ours (2M) & \best 1.00 & \best 0.96 & \best 0.82 & \best 1.00 & $0.97 \pm 0.10$ & \best 0.92 & \best 0.97 & \best 0.91 \\
Ours (200k) & \best 1.00 & 0.84 & 0.65 & \best 1.00 & \best $0.99 \pm 0.05$ & \second 0.76 & \second 0.96 & \second 0.90 \\
w/o $z \in \mathbb{S}^{31}$ & \best 1.00 & 0.80 & 0.64 & \best 1.00 & \best $0.99 \pm 0.05$ & 0.71 & 0.95 & 0.88 \\
w/o $\mathcal{L}_{\mathrm{con}}$ & \best 1.00 & \second 0.95 & \second 0.67 & 0.03 & $0.97 \pm 0.09$ & 0.59 & 0.82 & 0.73 \\
VAE & 0.44 & 0.06 & 0.33 & 0.01 & $0.93 \pm 0.13$ & 0.02 & 0.79 & 0.79 \\
\bottomrule
\end{tabular}
\end{table}

\subsection{Validity}
\label{app:ablation-validity}

We compared five models (Table~\ref{tab:latent-ablation}).
Ours (2M) is our full model trained on all 2M designs.
Ours (200k) is the same model trained on the 200k-design subset used for the
ablations and VAE, allowing a fair comparison (Sect.~\ref{sec:ablation}).
We encoded and decoded the 2{,}000 test designs with each model.
\textbf{App.\ acc.} is the appendage accuracy rate of designs decoded with the same number of appendages as the original,
and \textbf{Joint acc.} is the joint accuracy rate of designs decoded with the same number of joints.
\textbf{Geom. val.} is the fraction of decoded designs satisfying geometric validity with no overlapping
appendages and no geometry extending beyond the $128^3$ voxel grid.

Ours (2M) reconstructed designs best and had the highest geometric validity rate.
Most designs decoded from our latent genome passed the basic geometric checks,
though further checks are required before they can be built and simulated.
Among the ablations, training on reconstruction alone improved joint-count accuracy relative to ours (200k),
which is reasonable since reconstruction is its only objective.
The VAE performed poorly.
It recovered the number of appendages only for designs that happened to have five appendages,
almost never recovered the number of joints,
and produced
valid designs less than half of the time.
%

Despite higher joint accuracy and geometric validity rate than Ours (200k),
the reconstruction-only model performed poorly under CMA-ES
(Fig.~\ref{fig:fitness-curves}B), suggesting that genome smoothness also matters.

\begin{figure*}[!t]
  \centering
  \includegraphics[width=\columnwidth]{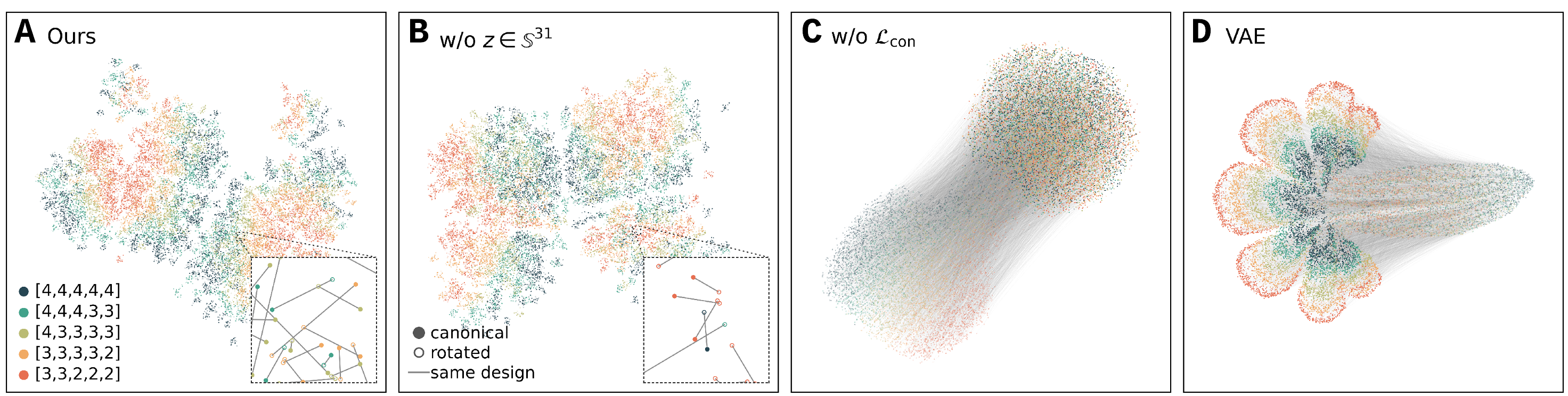}
  \caption{\textbf{Genetic coherence.}
  Ten thousand unique, pentadactyl morphologies were generated, copied, and the copy was rotated by a randomly selected angle.
  The resulting design pairs are visualized by UMAP
  \citep{healy2024uniform}
  along 2D projections of four different latent embeddings (\textbf{A-D}).
  A gray line connects the original (filled circle) to its rotated (open circle) counterpart.
  Five design variants were considered for this analysis,
  those with 
  a pair of three-jointed appendages and three two-jointed appendages (3,3,2,2,2),
  four three-jointed and one two-jointed (3,3,3,3,2),
  one four-jointed and four three-jointed (4,3,3,3,3),
  three four-jointed and 
  two three-jointed (4,4,4,3,3),
  and five four-jointed appendages (4,4,4,4,4).
  In the VAE, rotated copies can appear far from their originals in the UMAP projection (D).
  With the contrastive loss, rotated copies remain close to their originals
in the UMAP projection (A–B); without it, the two views separate (C).
  Removing the spherical projection, $z \in \mathbb{S}^{31}$, had little effect on this pairing (B).
  }
  \label{fig:latent-ablation}
\end{figure*}

%

\subsection{Smoothness}
\label{app:ablation-smoothness}

We assessed whether the genome is invariant to pose and whether genetic
distance reflects changes in body plan.

To test pose invariance, we encoded augmented copies of the test designs
using the methods in Fig.~\ref{fig:appx_augment}, excluding segment removal.
\textbf{Recall@1} is the fraction of augmented copies whose nearest neighbor
among the encoded originals of the 2{,}000 test designs is its own original.
Both Ours (2M) and Ours (200k), as well as the model without the spherical
projection, achieved 1.00.
The reconstruction-only model and the VAE achieved 0.03 and 0.01, respectively.

Fig.~\ref{fig:latent-ablation} visualizes one of these augmentations,
rotation about the vertical axis.
We generated 10{,}000 designs with five appendages from five topologies that differ by two segments from one to the next,
encoded each design before and
after rotating it about the vertical axis,
and projected the resulting genes into two dimensions using UMAP (\cite{healy2024uniform}).
Each design appears as two points joined by a line and colored by its topology.
With the contrastive loss (Fig.~\ref{fig:latent-ablation}A-B), the two points of a design 
appear close together
and designs of the same topology form patches that blend into those of neighboring topologies.
Without it (Fig.~\ref{fig:latent-ablation}C), the original and rotated
designs occupy separate regions.
In the VAE (Fig.~\ref{fig:latent-ablation}D), the original designs form a flower-like shape.
Designs with fewer joints (\textcolor{red}{red}) lie along the outer petals, while designs
with more joints (\textcolor{green!50!black}{green}) lie closer to the center.
After rotation, their points move away from the petals into the stem-like
region to the right.

We next tested whether genetic distance tracks controlled edits to a body plan.
For each test design, we removed up to four distal segments, one at a time,
and encoded each edited copy.
\textbf{Edit $\rho$} is the Spearman correlation between the number of
segments removed and the genetic distance from the original, averaged over
designs.
It was high for all five models (Table~\ref{tab:latent-ablation}), indicating
that genetic distance generally increased as more segments were removed.
We next measured whether independently sampled designs with nearby genes
have similar body plans.
We encoded a separate set of 4{,}500 designs from 17 topologies, with 900
designs for each number of appendages.
\textbf{Topology 10-NN acc.} is the fraction of designs whose topology matches
the most common topology among the decoded designs of their 10 nearest
genetic neighbors.
For each design in this set, we estimated total appendage length and body
volume by averaging the corresponding properties of its 10 nearest genetic
neighbors.
\textbf{Length 10-NN $r$} and \textbf{Volume 10-NN $r$} are the Pearson
correlations, across these 4{,}500 designs, between the estimates and the
true values.

Ours (2M) scored highest on all three neighborhood metrics, and the VAE
scored lowest.
Removing the contrastive loss improved joint accuracy but reduced
pose invariance and all three neighborhood metrics.
This pattern is consistent with the reconstruction-only model's poorer
CMA-ES performance (Fig.~\ref{fig:fitness-curves}B), suggesting that
coherent genetic neighborhoods support latent search beyond accurate
reconstruction.

\section{Hyperparameters and BOM}

\begin{table}[!htb]
\centering
\caption{Encoder architecture.}
\label{tab:encoder}
\begin{tabular}{|c|p{0.72\textwidth}|}
\hline
\textbf{Layer} & \textbf{Description} \\
\hline
Voxel input & $128^3 \times 2$ \\
\hline
Stem & Conv, kernel size $3^3$, stride 2, to $64^3$, 32 channels, batch normalization, ReLU \\
Stage 1 & 2$\times$ ResBlock, downsample to $32^3$, 64 channels \\
Stage 2 & 2$\times$ ResBlock, downsample to $16^3$, 128 channels \\
Stage 3 & 2$\times$ ResBlock, downsample to $8^3$, 256 channels \\
Stage 4 & 2$\times$ ResBlock, downsample to $4^3$, 512 channels \\
Average pool & out: 512 \\
\hline
ResBlock & 2$\times$ (Conv, kernel size $3^3$, batch normalization), skip connection, ReLU; 
           the first ResBlock of each stage downsamples with stride 2 \\
\hline
Graph input & up to 26 nodes (root and joints) $\times$ 6 (location, hinge axis) \\
\hline
Linear & in: 6, out: 128, ReLU \\
Message passing & 3$\times$ (mean over neighbors, Linear in: 256, out: 128, ReLU, skip connection) \\
Readout & mean and max over nodes, Linear in: 256, out: 128 \\
\hline
Fusion & in: 512 + 128, out: 512, batch normalization, ReLU; Linear in: 512, out: 512 \\
Projection & Linear in: 512, out: 32, normalized to unit length \\
\hline
Output & $z \in S^{31}$ \\
\hline
\end{tabular}
\end{table}


\begin{table}[!htb]
\centering
\caption{Decoder architecture.}
\label{tab:decoder}
\begin{tabular}{|c|p{0.72\textwidth}|}
\hline
\textbf{Layer} & \textbf{Description} \\
\hline
Input & $z \in S^{31}$, and the tokens predicted so far \\
\hline
Memory & Linear in: 32, out: $8 \times 384$ (eight vectors) \\
Embedding & token embedding (137) + learned position embedding (368), 384 channels \\
Layers 1--4 & Transformer decoder layer: causal self-attention, attention to the eight vectors, 
              feed-forward (1536, GELU); 8 heads, layer normalization before each part \\
Head & Linear in: 384, out: 137 \\
\hline
Output & logits of the next token: 137 \\
\hline
\end{tabular}
\end{table}

\begin{table}[h]
\centering
\caption{CMA-ES hyperparameters.}
\label{tab:cmaes_hyperparameters}
\begin{tabular}{|c|c|}
\hline
\textbf{Parameter} & \textbf{Value} \\
\hline
\multicolumn{2}{|l|}{\textit{CMA-ES}} \\
\hline
Population size & 32 \\
Initial step size ($\sigma_0$) & 0.25 \\
\hline
\multicolumn{2}{|l|}{\textit{Fitness of Rejected Designs}} \\
\hline
Failed to compile & $-1000$ \\
Motors do not fit & $-950$ \\
No initial pose found & $-900$ \\
Object dropped before training & $-850$ \\
\hline
\end{tabular}
\end{table}

\begin{table}[h]
\centering
\caption{Domain randomization used for fine-tuning.}
\label{tab:domain-randomization}
\begin{tabular}{|c|c|}
\hline
\textbf{Parameter} & \textbf{Range} \\
\hline
\multicolumn{2}{|l|}{\textit{Observation}} \\
\hline
Joint position noise (per step) & 0 to 0.02 rad \\
Joint position offset (per episode) & $\pm$0.005 rad \\
\hline
\multicolumn{2}{|l|}{\textit{Action}} \\
\hline
Action noise (per step) & 0.05 to 0.1 \\
Action delay & 0 to 2 control steps \\
\hline
\multicolumn{2}{|l|}{\textit{Initial State}} \\
\hline
Joint position & $\pm$0.02 rad \\
Joint velocity & $\pm$0.01 rad/s \\
Object position & $\pm$0.005 m \\
Object pitch and roll & $\pm$0.1 rad \\
\hline
\multicolumn{2}{|l|}{\textit{Design}} \\
\hline
Joint stiffness & 2.8 to 3.1 N\,m/rad \\
Joint damping & 0.08 to 0.12 N\,m\,s/rad \\
Restitution & 0 to 0.2 \\
\hline
\multicolumn{2}{|l|}{\textit{Object}} \\
\hline
Friction coefficient & 0.3 to 1.5 \\
Restitution & 0 to 0.2 \\
Mass scale & 0.9 to 1.1 \\
Size scale & 1.1 to 1.2 \\
\hline
\end{tabular}
\end{table}

\begin{table}[!t]
\centering
\caption{Reinforcement learning hyperparameters.}
\label{tab:ppo}
\begin{tabular}{|l|c|}
\hline
Parameter & Value \\
\hline
Parallel environments & 4096 \\
PPO iterations & 500 for rotate and flip, 200 otherwise \\
Steps per iteration & 32 \\
Policy and value network & GRU (256) + MLP (512, 256, 128), ELU \\
Learning rate & $5 \times 10^{-4}$ ($3 \times 10^{-4}$), adaptive \\
Target KL divergence & 0.02 \\
Discount ($\gamma$) & 0.99 \\
GAE ($\lambda$) & 0.95 \\
Clip range & 0.2 \\
Value loss weight & 4.0 (2.0) \\
Entropy weight & 0 (0.002) \\
Learning epochs per iteration & 5 \\
Mini-batches per epoch & 1 (4) \\
Gradient clipping & 1.0 \\
\hline
\end{tabular}
\end{table}

\begin{table}[!t]
\centering
\caption{Bill-of-materials (BOM) of the five physical designs. All designs
share the same parts: the base and every segment are 3D-printed in PLA, and
each actuated joint is one DYNAMIXEL XL330-M288-T that drives the child link
through an FPX330-H101 hinge frame and is seated in the parent link on an
FPX330-S102 (radial mount) or FPX330-S101 (lateral mount) frame. The first
column gives the unit cost (ROBOTIS US list price, September 2026) and the
remaining columns the quantity used by each design; PLA is counted by mass.
Totals are rounded to the nearest dollar. $^{\mathrm{a}}$Sold in sets of
four; unit cost is one quarter of the set price. $^{\mathrm{b}}$One cable
ships with every actuator, so cables add no cost. $^{\mathrm{c}}$M3 heat-set insert
cost is based on a 100-pack of Uxcell M3$\times$6\,mm brass heat-set inserts
(5\,mm OD) at \$12.69. PLA cost is estimated at \$0.02/g, with mass
estimated from the printed volume at the effective printed density.}
\label{tab:bom}
\vspace{4pt}
\begingroup
\footnotesize
\setlength{\tabcolsep}{6pt}
\renewcommand{\arraystretch}{1.15}
\begin{tabular}{|l|c|c|c|c|c|c|}
\hline
Item & Cost (US) & Rotate & Rotate-4 & Pick & Hang & Retrieve \\
\hline
XL330-M288-T Actuator & \$27.49 & 15 & 15 & 10 & 10 & 9 \\
\hline
FPX330-H101 Hinge Frame with Idler$^{\mathrm{a}}$ & \$2.82 & 15 & 15 & 10 & 10 & 9 \\
\hline
FPX330-S102 Base Frame (Radial)$^{\mathrm{a}}$ & \$2.50 & 7 & 7 & 10 & 5 & 1 \\
\hline
FPX330-S101 Side Frame (Lateral)$^{\mathrm{a}}$ & \$2.50 & 8 & 8 & 0 & 5 & 8 \\
\hline
Robot Cable-X3P 180\,mm$^{\mathrm{b}}$ & incl. & 15 & 15 & 10 & 10 & 9 \\
\hline
U2D2 USB Converter & \$36.92 & 1 & 1 & 1 & 1 & 1 \\
\hline
U2D2 Power Hub Board Set & \$21.85 & 1 & 1 & 1 & 1 & 1 \\
\hline
M3 Heat-Set Insert$^{\mathrm{c}}$ & \$0.13 & 180 & 180 & 120 & 120 & 108 \\
\hline
PLA Filament (g)$^{\mathrm{c}}$ & \$0.02 & 223 & 245 & 219 & 103 & 99 \\
\hline
\textbf{Total} & & \textbf{\$578} & \textbf{\$579} & \textbf{\$406} & \textbf{\$404} & \textbf{\$370} \\
\hline
\end{tabular}
\endgroup
\end{table}

\clearpage

\clearpage
\section*{Generative AI Usage}


We used generative AI tools to assist in implementing methods, creating figures, and editing some parts of the Appendix for readability.
For all other tasks,
generative AI was not used. 
We reviewed all AI-assisted contributions for correctness and take responsibility for the final contents of this paper.

\end{document}